\documentclass[pre,twocolumn,twoside,byrevtex,superscriptaddress,floatfix,nofootinbib]{revtex4-1}

\usepackage[realmainfile]{currfile}

\usepackage{color}

\usepackage{graphicx,epsfig,verbatim,enumerate}
\usepackage{amssymb,amsmath}
\usepackage{ifthen}

\usepackage{longtable}

\usepackage{mathtools}

\newboolean{twocolswitch}

\newcommand{\sindex}[1]{}
\newcommand{\nindex}[1]{}

\newcommand{\www}[1]{\url{#1}}

\usepackage{lettrine}

\usepackage{hyperref}
\urldef\onlineappendixurl
\url{http://compstorylab.org/stretchablewords/}

\usepackage{alltt}

\usepackage{verbatim}

\newcommand{\wordtrees}{spelling trees }
\newcommand{\Wordtrees}{Spelling trees }
\newcommand{\wordtree}{spelling tree }
\newcommand{\Wordtree}{Spelling tree }
\newcommand{\balanceplots}{balance plots }

\setboolean{twocolswitch}{true}

\begin{document}

\title{\protect
  Hahahahaha, Duuuuude, Yeeessss!:  A two-parameter characterization of stretchable words and the dynamics of mistypings and misspellings
}

\author{
\firstname{Tyler J.}
\surname{Gray}
}
\email{tyler.gray@uvm.edu}

\affiliation{
  Vermont Complex Systems Center,
  Computational Story Lab,
  Department of Mathematics \& Statistics,
  The University of Vermont,
  Burlington, VT 05401.
  }

\author{
\firstname{Christopher M.}
\surname{Danforth}
}
\email{chris.danforth@uvm.edu}

\affiliation{
  Vermont Complex Systems Center,
  Computational Story Lab,
  Department of Mathematics \& Statistics,
  The University of Vermont,
  Burlington, VT 05401.
  }

\author{
\firstname{Peter Sheridan}
\surname{Dodds}
}
\email{peter.dodds@uvm.edu}

\affiliation{
  Vermont Complex Systems Center,
  Computational Story Lab,
  Department of Mathematics \& Statistics,
  The University of Vermont,
  Burlington, VT 05401.
  }

\date{\today}

\begin{abstract}
  \protect
  Stretched words like `heellllp' or `heyyyyy' are a regular feature of spoken language, often used to emphasize or exaggerate the underlying meaning of the root word. While stretched words are rarely found in formal written language and dictionaries, they are prevalent within social media.  In this paper, we examine the frequency distributions of `stretchable words' found in roughly 100 billion tweets authored over an 8 year period. We introduce two central parameters, `balance' and `stretch',  that capture their main characteristics, and explore their dynamics by creating visual tools we call `\balanceplots\unskip' and `\wordtrees\unskip'.  We discuss how the tools and methods we develop here could be used to study the statistical patterns of mistypings and misspellings, along with the potential applications in augmenting dictionaries, improving language processing, and in any area where sequence construction matters, such as genetics.

\end{abstract}


\maketitle

\section{Introduction}
\label{sec:introduction}

Watch a soccer match, and you are likely to hear an announcer shout `GOOOOOOOOOAAAAAAAAL!!!!!!'.  Stretched words, sometimes called elongated words \cite{elongated}, are an integral part of spoken language, often used to modify the meaning of the base word in some way, such as to strengthen the meaning (e.g., `huuuuuge'), imply sarcasm (e.g., `suuuuure'), show excitement (e.g., `yeeeessss'), or communicate danger (e.g., `nooooooooooooo').  We will refer to words that are amenable to such lengthening as `stretchable words'.    

However, despite their being a fundamental part of spoken language, stretched words are rarely found in literature and lexicons:  There is no `hahahahahahaha' in the Oxford English Dictionary \cite{OED1989}.  
With the advent and rise of social media, stretched words have finally found their way into large-scale written text.  

With the increased use of social media comes rich datasets of a linguistic nature, granting science an unprecedented opportunity to study the everyday linguistic patterns of society.  As such, in recent years there have been a number of papers published that have used data from social media platforms, such as Twitter, to study different aspects of language \cite{grieve2016a, gray2018a, goncalves2014a, goncalves2018a, donoso2017a, eisenstein2010a,  eisenstein2014a}.  

In this paper, we use an extensive set of social media messages collected from Twitter---tweets---to investigate the characteristics of stretchable words used in this particular form of written language.  
The tools and approach we introduce here have many potential applications, including the possible use by dictionaries to formally include this intrinsic part of language.  The online dictionary Wiktionary has already discussed the inclusion of some stretched words and made a policy on what to include \cite{wiktionary:talk_cute, wiktionary:criteria_for_inclusion}.  
Other potential applications include the use by natural language processing software and toolkits, and by Twitter to build better spam filters. 

We structure our paper as follows:  In Sec.~\ref{sec:data}, we detail our dataset and our method of collecting stretchable words and distilling them down to their `kernels'.  In Sec.~\ref{sec:distributions}, we examine the frequency distributions for lengths of stretchable words.  We quantify two independent properties of stretchable words:  Their `balance' in Sec.~\ref{sec:balance} and `stretch' in Sec.~\ref{sec:stretch}.  In Sec.~\ref{sec:trees}, we develop an investigative tool, `\wordtrees\unskip', as a means of visualizing stretchable words involving a two character repeated element.  We comment on mistypings and misspellings in Sec.~\ref{sec:mistypings}.  Finally, in Sec.~\ref{sec:concludingremarks}, we provide concluding remarks.  

\section{Description of the dataset and method for extracting stretched words}
\label{sec:data}

\begin{table}[htbp!]
\begin{center}
\begin{tabular}{|r|p{.7\columnwidth}|}
\hline
1. & hahhahahaahahaa \newline $\rightarrow$ (ha) \\
\hline
2. & gooooooaaaaaaal \newline $\rightarrow$ g[o][a]l \\
\hline
3. & ggggoooooaaaaallllll \newline $\rightarrow$ [g][o]aaaaallllll \newline $\rightarrow$ [g][o][a][l] \\
\hline
4. & bbbbbaaaaaabbbbbbyyyyyyy \newline $\rightarrow$ [b][a][b]yyyyyyy \newline $\rightarrow$ [b][a][b][y] \\
\hline
5. & awawawaaawwwwwesssssommmmmeeeeee \newline $\rightarrow$ (aw)esssssommmmmeeeeee \newline $\rightarrow$ (aw)essssso[m][e] \newline $\rightarrow$ (aw)e[s]o[m][e] \\
\hline 
\end{tabular}
\end{center}
\caption{Examples of distilling tokens down to their kernels.  The first line of each cell is the example token.  The following lines show the result after every time a replacement of characters by the corresponding single letter element(s) or double letter element is made by the code, in order.  The final line of each cell gives the resulting kernel for each example.  }\label{table:kernel_creation}
\end{table}

The Twitter dataset we use in this study comprises a random sample of approximately 10\% of all tweets (the `gardenhose' API) from 9 September 2008 to 31 December 2016.  We limited our scope to tweets that either were flagged as an English tweet or not flagged for any language.  All tweets in this time period have a maximum length of 140 characters.  To collect stretchable words, we begin by making all text lowercase and collecting all tokens within our dataset from calendar year 2016 that match the Python regular expression r`(\textbackslash b\textbackslash w*(\textbackslash w)(\textbackslash w)(?:\textbackslash 2$|$\textbackslash 3)\{28,\}\textbackslash w*\textbackslash b)'.  This pattern will collect any token with at least 30 characters that has a single character repeated at least 29 times consecutively, or two different characters that are repeated in any order at least 28 times, for a total of at least 30 consecutive repeated occurrences of the two characters. 
The choice of 28 in the regular expression is a threshold we chose with the goal of limiting our collection to tokens of words that really do get stretched in practice.

After collecting these tokens, we remove any that contains a character that is not a letter ([a-z]), and distill each remaining token down to its `kernel'.  Table~\ref{table:kernel_creation} gives a few examples of this distillation process.  Proceeding along the token from left to right, whenever any pair of distinct letters, $l_1$ and $l_2$, occur in the token where 1.~$l_1$ occurs followed by any sequence of $l_1$ and $l_2$ of total length at least three, and 2.~such that $l_1$ and $l_2$ each occur at least twice in the sequence, we replace the sequence with the `two letter element' $(l_1l_2)$.  For example, see the first cell in Table~\ref{table:kernel_creation}. 

 Exceptions to the preceding are: 1.~The case where the sequence is a series of $l_1$ followed by a series of $l_2$, which is replaced with the pair of `single letter elements' $[l_1][l_2]$.  For example, see the second cell in Table~\ref{table:kernel_creation}.  
And 2., the case where the sequence is a series of $l_1$ followed by a series of $l_2$ followed by a series of $l_1$, which is replaced with $[l_1][l_2][l_1]$.  For example, see the first step in the fourth cell of Table~\ref{table:kernel_creation} where `bbbbbaaaaaabbbbbb' is replaced with [b][a][b].  

Following this process, whenever a single letter, $l_3$, occurs two or more times in a row, we replace the sequence with the single letter element $[l_3]$.  For example, see the last step of the fourth cell in Table~\ref{table:kernel_creation} where `yyyyyyy' is replaced with [y], or the last step in the fifth cell where `sssss' is replaced with [s]. 

\begin{figure}[tp!]
\centering
\includegraphics[width=\columnwidth]{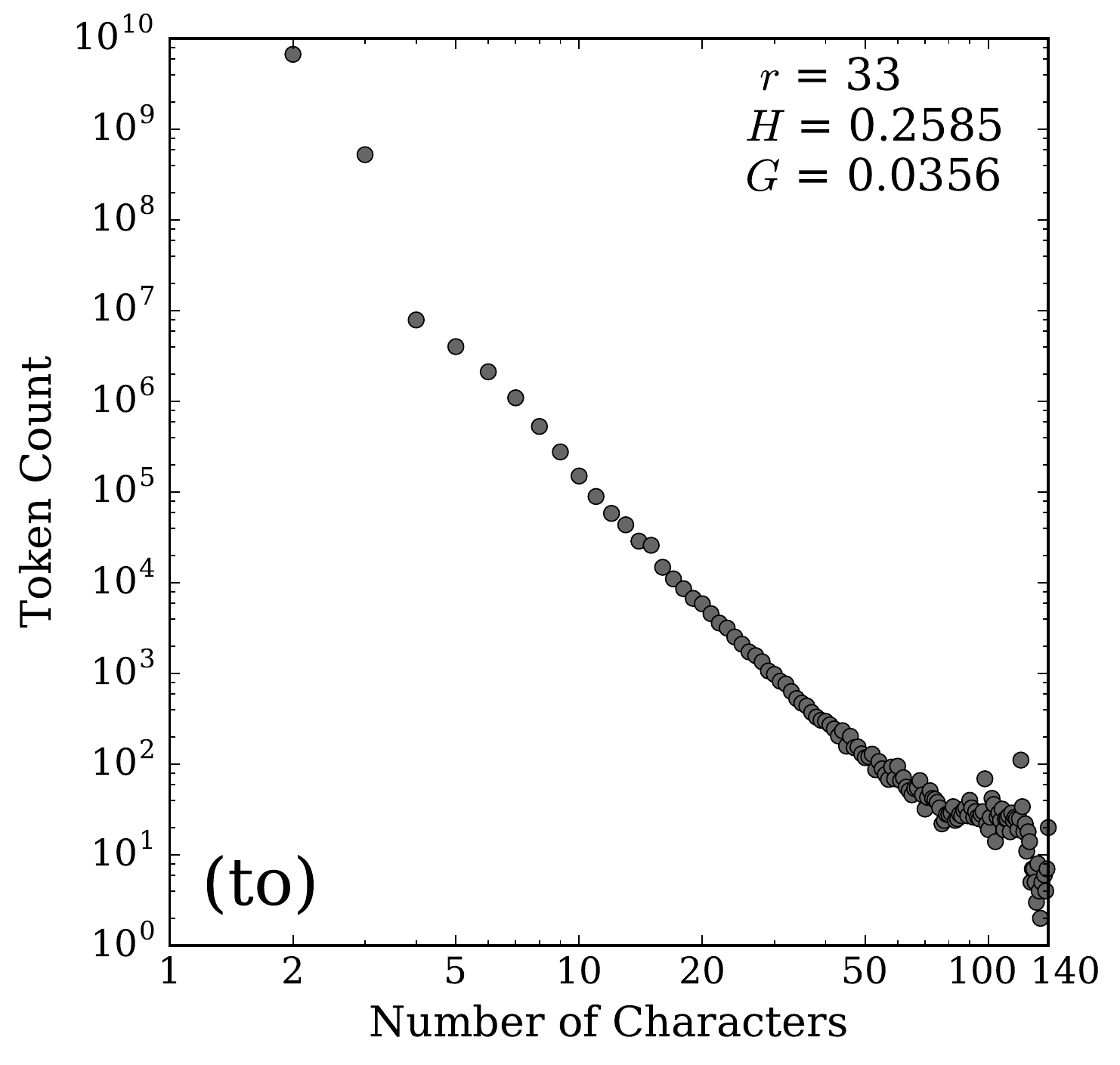}
\caption{Token count distribution for the kernel (to).  The horizontal axis represents the length (number of characters) of the token and the vertical axis gives the total number of tokens of a given length that match this kernel.  The included statistics give the kernel rank, $r$ (see Sec.~\ref{sec:data}), the value of the balance parameter (normalized entropy, $H$; see Sec.~\ref{sec:balance}),  and the value of the stretch parameter (Gini coefficient, $G$; see Sec.~\ref{sec:stretch}) for this kernel.  The large drop between the second and third points denotes the change from `unstretched' versions of (to), located to the left of this drop, to `stretched' versions of (to), located to the right of this drop.}
\label{fig:to-powerlaw}
\end{figure}

\begin{figure}[tp!]
\centering
\includegraphics[width=\columnwidth]{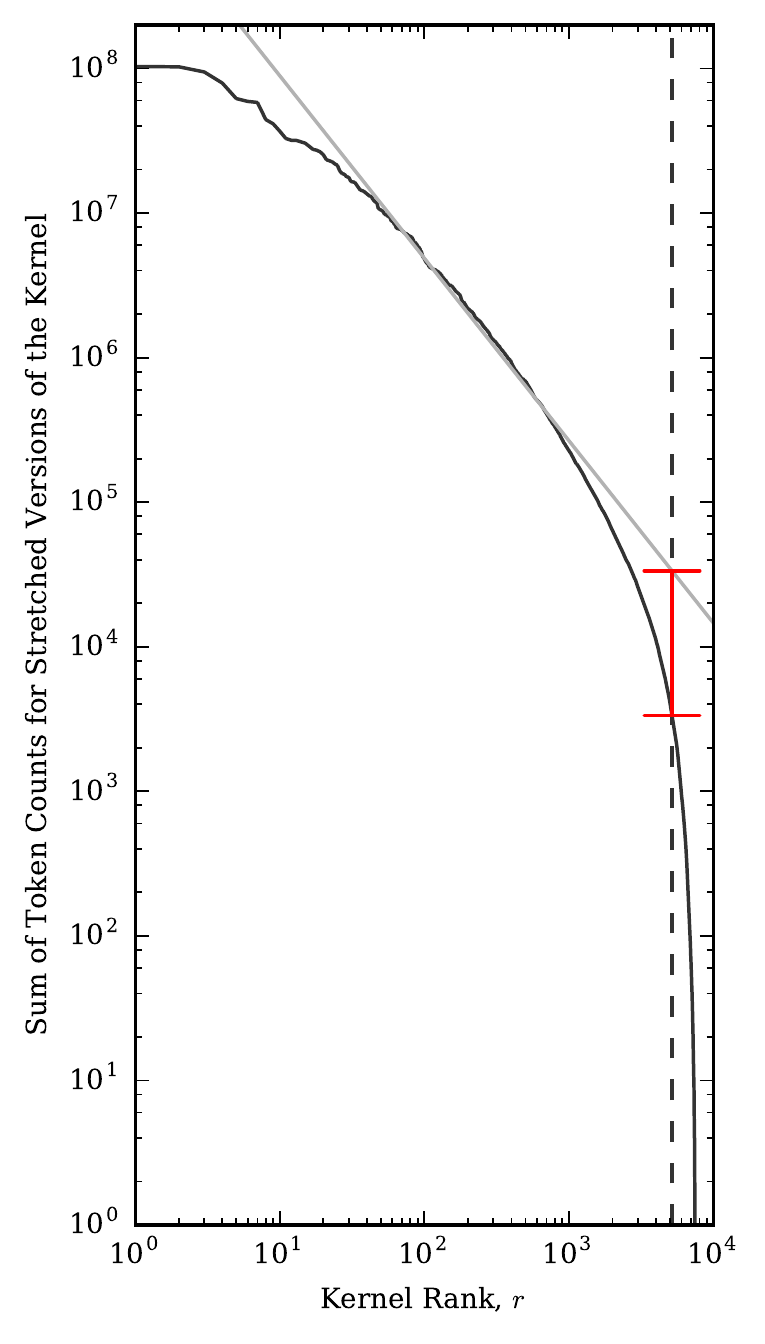}
\caption{Total token counts for stretched versions of all kernels.  Kernels are ranked by their descending total token count along the horizontal axis.  The diagonal line gives the regression line calculated using the values between ranks 10 and $10^3$.  The vertical dashed line denotes the first location after rank $10^3$ where the distribution drops below $1/10$ of the corresponding value of the regression line, denoted by the red interval, giving the cutoff rank for the final threshold to decide which kernels to include in the study. } 
\label{fig:zipf_distribution}
\end{figure}

We collected tokens in batches of seven consecutive days at a time throughout 2016 (with the last batch being only two days).  If a kernel is not found in more than one batch, or within the same batch but from at least two distinct stretched words, then it is removed from consideration.

Different but related stretched words (that is, different stretched words, but both stretched out versions of the same base word) may distill to different kernels.  
We combine these into a single kernel for each word such that it covers all cases observed in the collected tokens.  For example, the kernels g[o]a[l] and go[a][l] would be combined as g[o][a][l].  The kernels h[a] and (ha) would be combined as (ha).

After processing our dataset, we obtained a collection of 7,526 kernels.  
We then represented each kernel with a corresponding regular expression and collected all tokens in our entire gardenhose dataset that matched the regular expressions. To go from the kernel to the regular expression, we replaced ] with ]+, replaced $(l_1l_2)$ with $l_1[l_1l_2]\text{*}l_2[l_1l_2]\text{*}$, and we surrounded the kernel with word boundary characters \textbackslash b.  So, for example, the kernel g[o][a][l] goes to the Python regular expression r`\textbackslash bg[o]+[a]+[l]+\textbackslash b' and the kernel (ha) goes to the Python regular expression r`\textbackslash bh[ha]*a[ha]*\textbackslash b'.  

Once we collected all tokens matching our kernels, we carried out a final round of thresholding on our kernel list, removing those with the least amount of data and least likely to represent a bona fide stretchable word.  For each kernel, we calculated the token count as a function of token length (number of letters) for all tokens matching that kernel.  For example, Fig.~\ref{fig:to-powerlaw} gives the plot of the token count distribution for the kernel (to).  Then, with the token counts in order by increasing token length, as in Fig.~\ref{fig:to-powerlaw}, we found the location of the largest drop in the $\log_{10}$ of token counts between two consecutive values within the first 10 values.  We call the words with lengths coming before the location of the drop `unstretched' versions of the kernel and those that come after `stretched' versions.  For most kernels, the largest drop will be between the first and second value.  However, for some kernels this drop occurs later.  For example, in Fig.~\ref{fig:to-powerlaw} we see that for the kernel (to), which covers both the common words `to' and `too', this drop is between the second and third value (between tokens of length three and four).  Thus, the unstretched versions of (to) are represented by the first two points in Fig.~\ref{fig:to-powerlaw}, with the remaining points representing stretched versions of (to). 

We then ranked the kernels by the sum of the token counts for their stretched versions.  Fig.~\ref{fig:zipf_distribution} shows this sum as a function of rank for each kernel.  Inspired by the idea of a cutoff frequency \cite{wiki:cutoff_frequency}, we estimate a cutoff rank for the kernels.  Using the values between rank 10 and $10^3$, we found the regression line between the $\log_{10}$ of the ranks and the $\log_{10}$ of the summed token counts (straight line, Fig.~\ref{fig:zipf_distribution}).  We calculated the cutoff as the first rank (after $10^3$) where the summed token count is less than $1/10$ of the corresponding value of the regression line.  This occurs at rank 5,164, which is shown by the vertical dashed line in Fig.~\ref{fig:zipf_distribution}. For the remainder of this study, we used the kernels with rank preceding this cutoff, giving us a total of 5,163 kernels, and, unless otherwise specified, a kernel's `rank', $r$, refers to the rank found here.
See Online Appendix A at \onlineappendixurl \ for a full list of kernels meeting our thresholds, along with their regular expressions and other statistics discussed throughout the remainder of this paper.  

\section{Analysis and Results}
\label{sec:results}

\subsection{Distributions}
\label{sec:distributions}

\begin{figure}[tp!]
\centering
\includegraphics[width=\columnwidth]{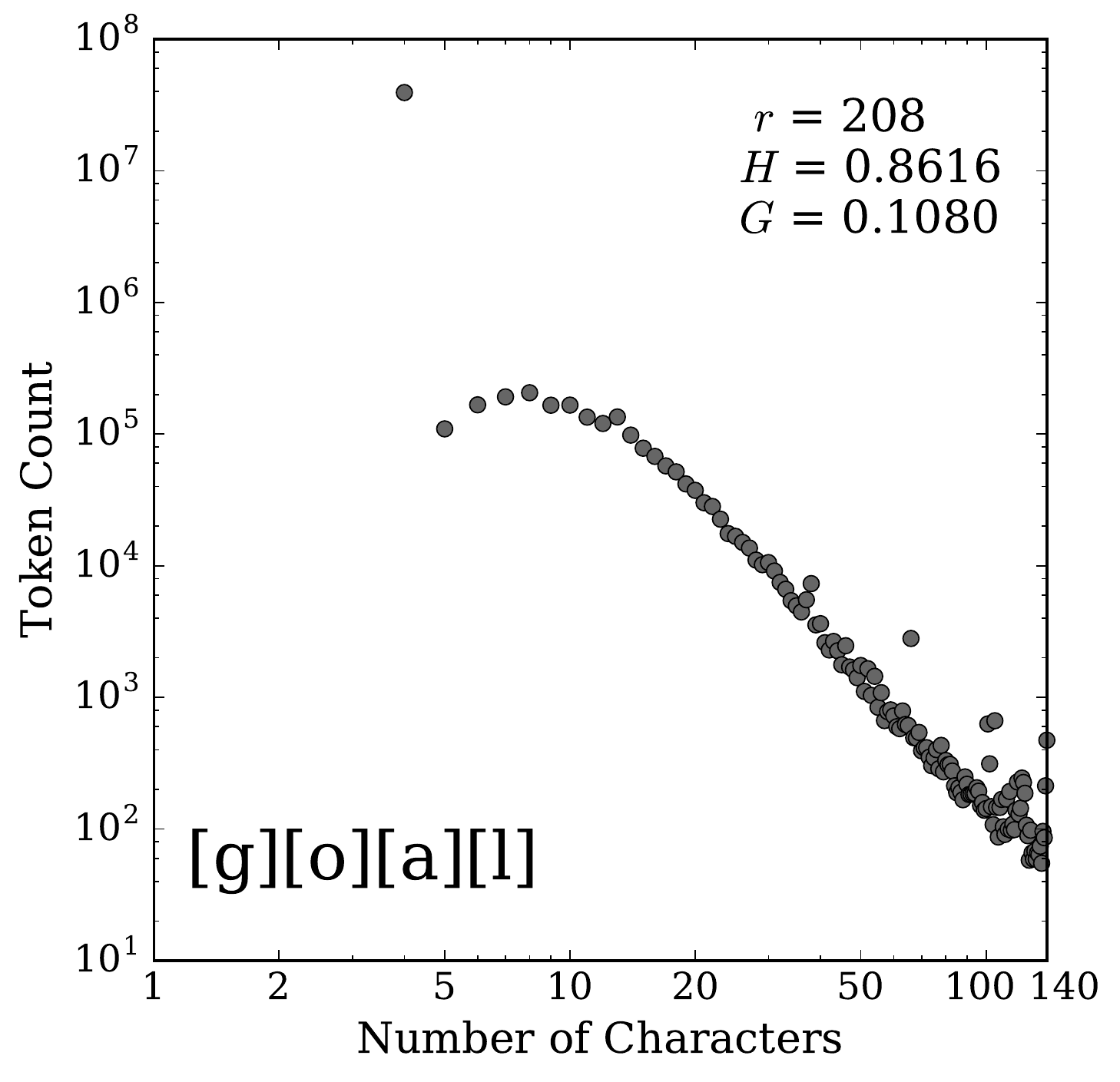}
\caption{Token count distribution for the kernel [g][o][a][l].  The horizontal axis represents the length (number of characters) of the token and the vertical axis gives the total number of tokens of a given length that match this kernel.  See Fig.~\ref{fig:to-powerlaw} caption for details on the included statistics. The base version of the word appears roughly 100 times more frequently than the most common stretched version.}
\label{fig:goal-powerlaw}
\end{figure}

For each kernel, we plotted the corresponding distribution of token counts as a function of token length.  Most of the distributions largely follow a roughly power-law shape.
For example, Fig.~\ref{fig:goal-powerlaw} gives the frequency distribution for the kernel [g][o][a][l].  From the elevated frequency of the first dot, we can see that the unstretched word `goal' is used about two orders of magnitude more frequently than any stretched version.  After the first point, we see a rollover in the distribution, showing that if users are going to stretch the word, they are more likely to include a few extra characters rather than just one.  We also see that there are some users who indeed fill the 140 character limit with a stretched version of the word `goal', and the elevated dot there suggests that if users get close to the character limit, they are more likely to fill the available space.  The other dots elevated above the trend represent tokens that likely appear in tweets that have a small amount of other text at the beginning or end, such as a player name or team name, or, more generally, a link or a user handle.

\begin{figure}[tp!]
\centering
\includegraphics[width=\columnwidth]{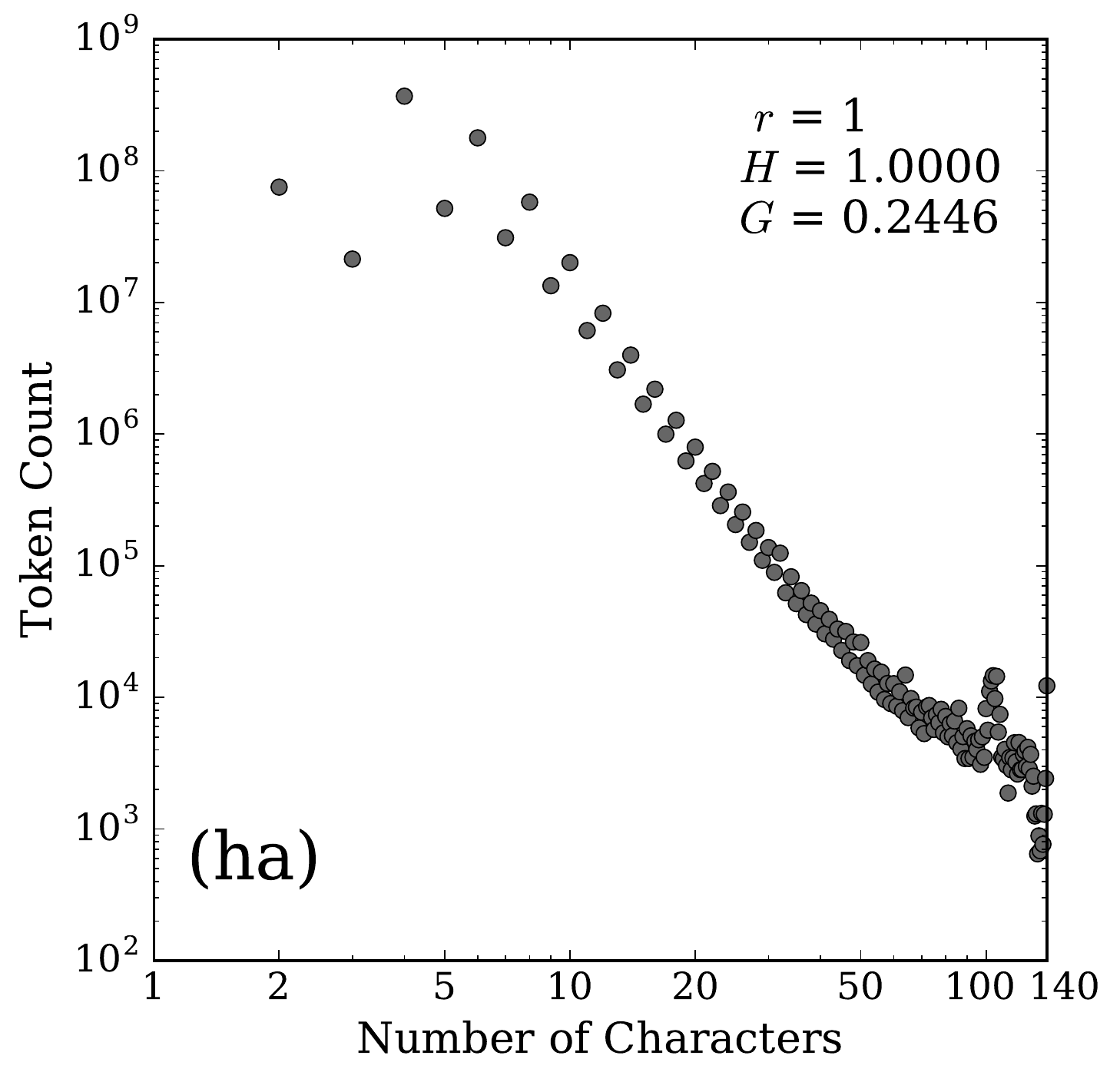}
\caption{Token count distribution for the kernel (ha).  The horizontal axis represents the length (number of characters) of the token and the vertical axis gives the total number of tokens of a given length that match this kernel.  See Fig.~\ref{fig:to-powerlaw} caption for details on the included statistics. }
\label{fig:ha-powerlaw}
\end{figure}

In Fig.~\ref{fig:ha-powerlaw}, we show the frequency distribution for the kernel (ha) as an example of a distribution for a two character repeated element.  For this distribution we observe an alternating up and down in frequency for even length tokens and odd length tokens.  This behaviour is typical of distributions with a two character repeated element, likely resulting from an intent for these tokens to be a perfect alternating repetition of `h' and `a', hahaha\dots, to represent laughter.  Under this assumption, the correct versions will be even length.  Then, any incorrect version could be odd or even length depending on the number of mistakes.  We look at mistakes further in Sec.~\ref{sec:mistypings}.  

We note that there is also an initial rollover in this distribution, showing that the four character token, with dominant contributor `haha', is the most common version for this kernel.  
We also again see some elevated counts near the tail, including for 140 characters, along with some depressed counts just short of 140, which again suggests that when users approach the character limit with stretched versions of (ha), they will most likely fill the remaining space.     
We did not perform a detailed analysis of this area, but it is likely that the other elevated points near the end are again due to the inclusion of a link or user handle, etc.  Similarly, the general flattening of the distribution's right tail is likely a result of random lengths of short other text combined with a stretched word that fills the remaining space.  

Similar distributions for each kernel can be found in Online Appendix B at \onlineappendixurl.

\subsection{Balance}
\label{sec:balance}

\begin{figure}[tp!]
\centering
\includegraphics[width=\columnwidth]{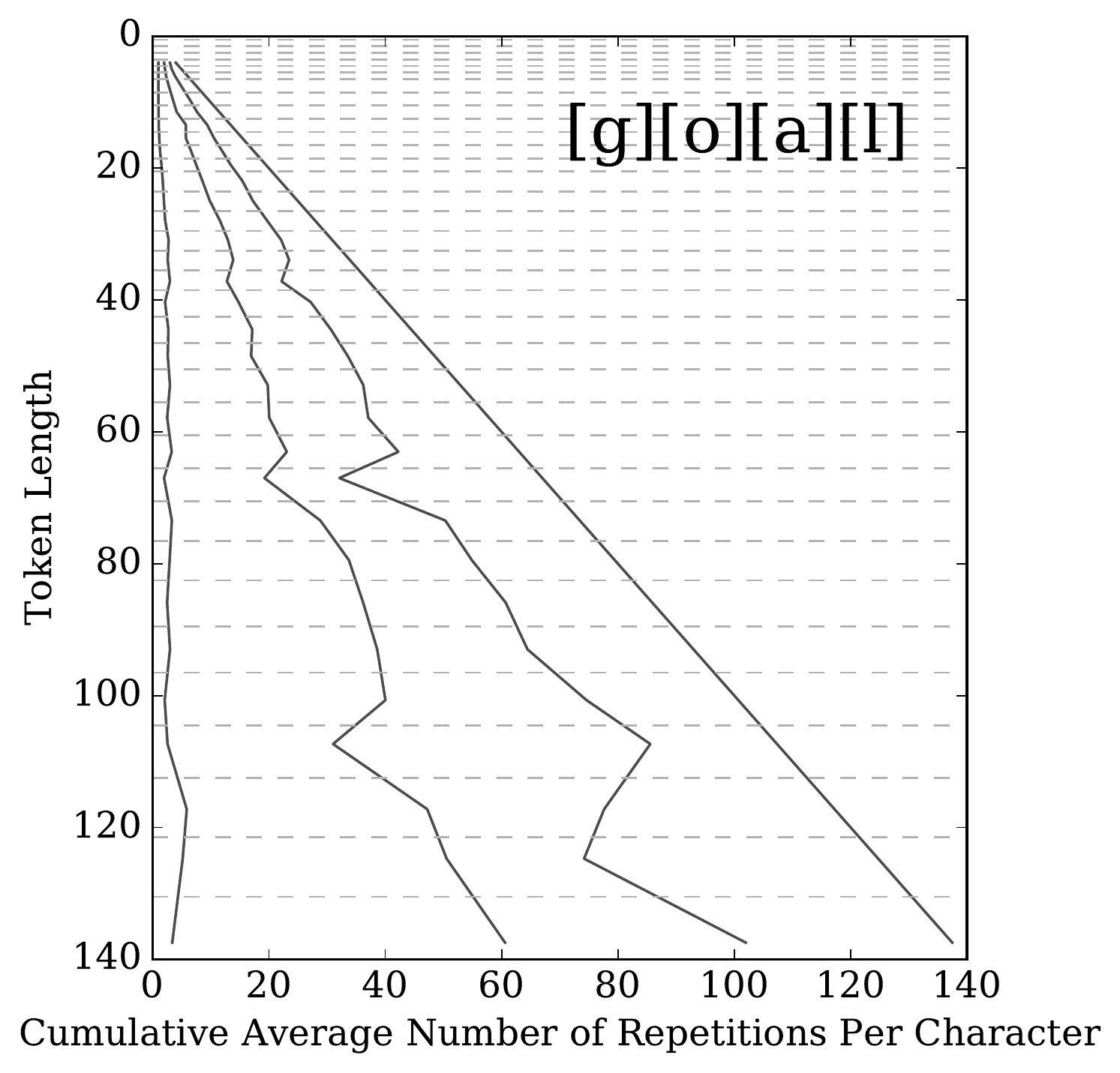}
\caption{Balance plot for the kernel [g][o][a][l].  The vertical axis represents the length (number of characters) of tokens, and is broken into bins of lengths, with boundaries denoted by horizontal dashed lines, which increase in size logarithmically.  For all the tokens that match the kernel and fall within a bin of lengths, the average number of times each character was stretched in those tokens was calculated, and is shown on the plot as the distance between two solid lines in the same order as in the kernel.  Thus, for a given bin, the distance between the vertical axis and the first solid line is the average stretch for the letter `g', the distance between that first line and the second line is the average stretch for the letter `o', and so on. For example, the last bin contains tokens with lengths in the interval $[ \protect131, \protect140 ]$, with average length roughly \protect137.  On average, tokens falling in this most celebratory bin contain roughly \protect3 `g's, \protect57 `o's, \protect41 `a's, and \protect36 `l's.}
\label{fig:goal-balance}
\end{figure}

For each kernel, we measure two quantities:  1.~The balance of the stretchiness across characters, and 2.~the overall stretchiness of the kernel.  To measure balance, we calculate the average stretch of each character in the kernel across all the tokens within a bin of token lengths.  By average stretch of a character, we mean the average number of times that character appears.  Fig.~\ref{fig:goal-balance} shows the balance for the kernel [g][o][a][l] partitioned into bins of logarithmically increasing sizes of length.  The horizontal dashed lines represent the bin edges.  The distance between the solid diagonal lines represents the average stretch, or average number of times each character was repeated, and are plotted in the same order that they appear in the kernel.  From this figure we see that `g' is not stretched much on average, `o' is stretched the most, and `a' and `l' are both stretched around 2/3 as much as `o'.  

\begin{figure}[tp!]
\centering
\includegraphics[width=\columnwidth]{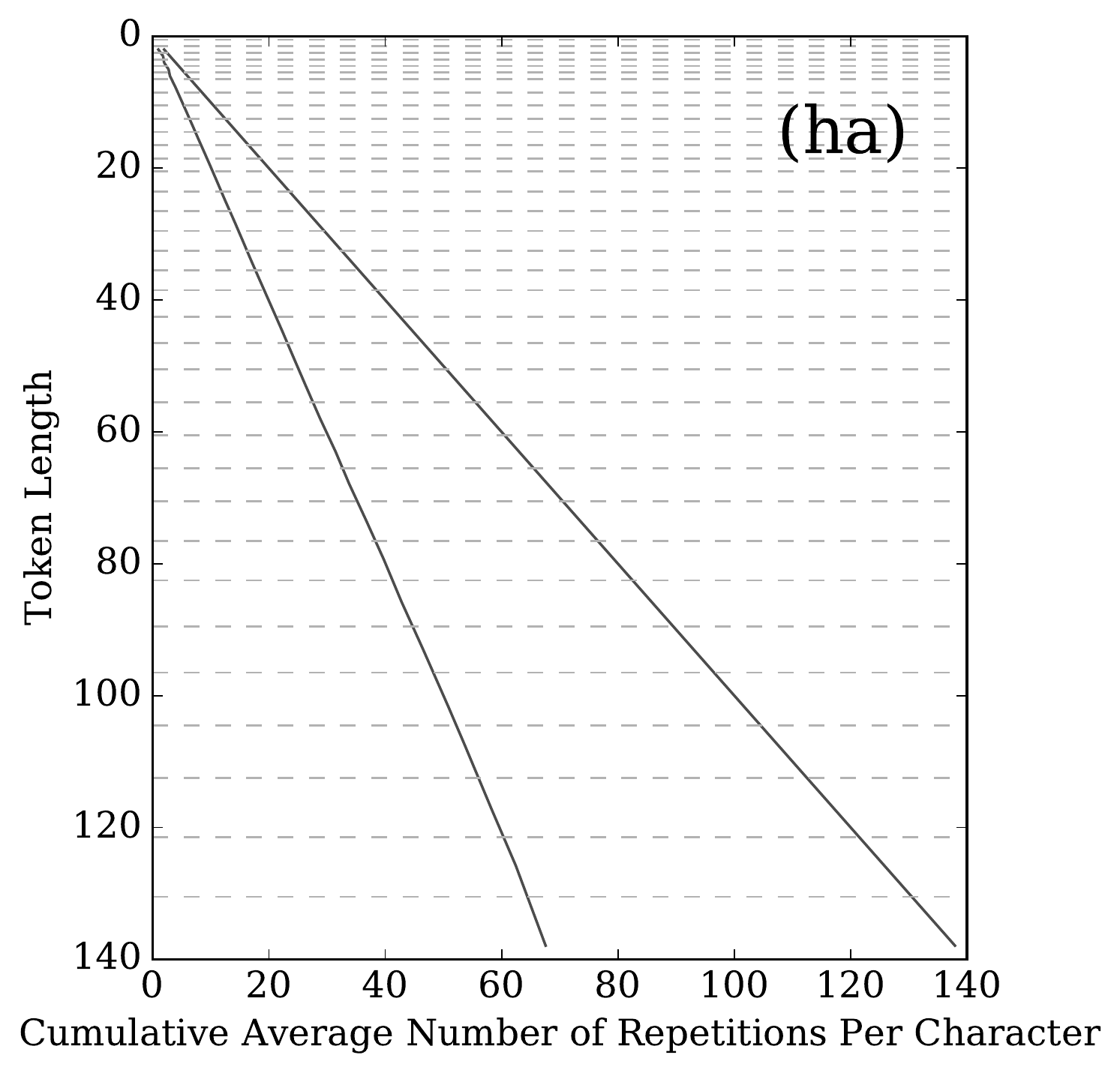}
\caption{Balance plot for the kernel (ha).  See the Fig.~\ref{fig:goal-balance} caption for plot details.  For two letter elements, even though the letters can alternate within a given token, we still count the number of occurrences for each letter separately and display the average number of total repetitions in the same order as the letters appear in the kernel.  Thus, for a given bin, the distance between the vertical axis and the first line is the average number of times the letter `h' occurred in the tokens, and the distance between that first line and the second line is the average number of times the letter `a' occurred in the token.  This plot clearly shows that (ha) is well balanced across all bins of token lengths.  }
\label{fig:ha-balance}
\end{figure}

When part of the kernel is a two letter element of the form $(l_1l_2)$, we still count the number of occurrences of $l_1$ and $l_2$ corresponding to this element in the kernel separately, even though the letters can be intermingled in the stretched word.  When we display the results, we display it in the same order that the letters appear in the kernel.  So in Fig.~\ref{fig:ha-balance}, which shows the results for the kernel (ha), the first space represents the average stretch for `h' and the second space is for `a'.  From this figure, we can see that the stretch is almost perfectly balanced between the two letters on average.  

\begin{figure}[tp!]
\centering
\includegraphics[width=\columnwidth]{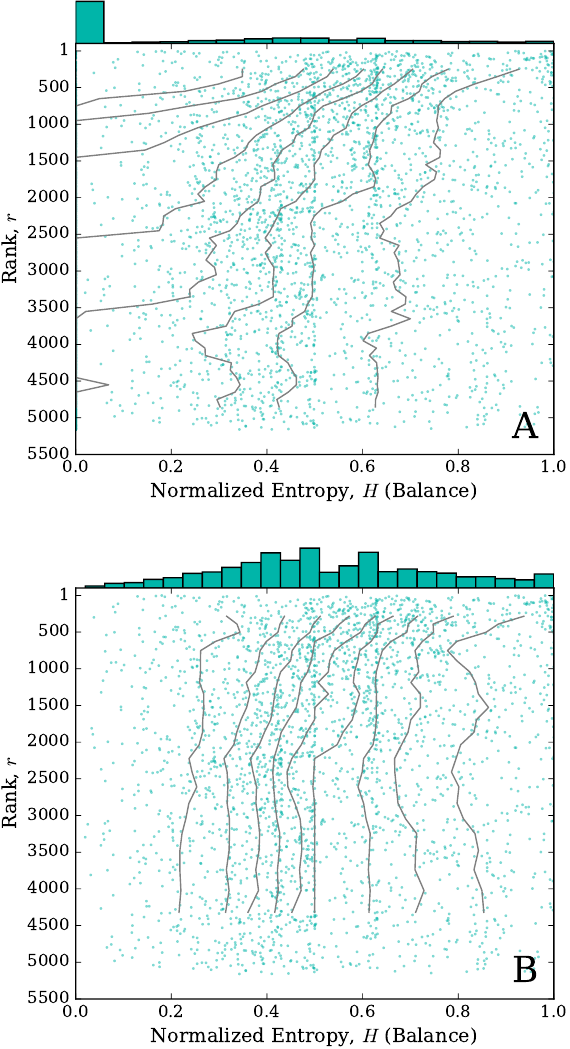}
\caption{Jellyfish plots for kernel balance for (A) all kernels, and (B) excluding kernels with entropy exactly 0.  Corresponding histograms are given at the top of each plot.  Kernels are plotted vertically by their rank, $r$, and horizontally by their balance as given by normalized entropy, $H$, where larger entropy denotes increased balance.  The deciles $0.1, 0.2, \dots, 0.9$ are calculated for rolling bins of 500 kernels and are plotted as the `tentacles'. }
\label{fig:balance_jellyfish}
\end{figure}

Similar balance plots can be found for each kernel in Online Appendix C at \onlineappendixurl.  In general, for these balance plots, we stop plotting at the first bin with no tokens, even if later bins may be nonempty.  

For each kernel, we also calculate an overall measure of balance.  
To do this, we begin by binning the tokens by length.  Then, for each bin (containing tokens longer than the kernel) we calculate the average stretch for each character across tokens within the bin as before.  
Then, we subtract one from each of these values (removing the contribution from each base character; counting just the number of times each character was repeated) and normalize the values so they sum to 1 and can be thought of like probabilities.  We then average the probabilities across the bins, weighing each bin equally, and compute the normalized entropy, $H$, of the averaged probabilities as our overall measure of balance.
This measure is such that if each character stretches the same on average, the normalized entropy is 1, and if only one character in the kernel stretches, the normalized entropy is 0.  Thus, higher entropy corresponds with more balanced words.  
(For a comparison with an alternate entropy measure where tokens contribute equally rather than equally weighing each length bin, and an explanation of the different corresponding views, see Appendix~\ref{appendix:entropy_alternate}.)

Fig.~\ref{fig:balance_jellyfish} shows two `jellyfish plots' \cite{kloumann2012a} for balance.  Fig.~\ref{fig:balance_jellyfish}A is the version containing all words and for Fig.~\ref{fig:balance_jellyfish}B we remove the words that have a value of 0 for entropy.  The top of the left plot in Fig.~\ref{fig:balance_jellyfish} shows the frequency histogram of the normalized entropy for each kernel.  
The spike containing value 0 comes largely from kernels where only one character stretches, giving that kernel an entropy of exactly 0.  
The main plot shows the normalized entropy values as a function of word rank, where rank is given, as before, by the sum of stretched token counts.  The `tentacles' give rolling deciles.  That is, for rolling bands of 500 words by rank, the deciles $0.1, 0.2, \dots, 0.9$ are calculated for the entropy values, and are represented by the solid lines.  These plots allow us to see how stable the distribution is across word ranks. 

We can see from Fig.~\ref{fig:balance_jellyfish}A that the distribution largely shifts towards smaller entropy values with increasing rank, mostly drawn in that direction by the kernels with only a single repeated letter and entropy exactly 0.  
For Fig.~\ref{fig:balance_jellyfish}B, we remove all kernels with entropy 0.  Everything else remains the same, including the rank of each kernel (we skip over ranks of removed kernels) and the rolling bands of 500 kernels for percentile calculations still have 500 kernels, and thus tend to be visually wider bands.  
In contrast to Fig.~\ref{fig:balance_jellyfish}A, we now see a small left-shift in the earlier ranks, and then the distribution tends to stabilize for lower ranks.  This shows that the highest ranked kernels tend to have a larger entropy, meaning the stretch of the kernel is more equally balanced across all characters.  We also see that not many of the high ranked words stretch with just one character.  It appears that these kernels that stretch in only a single character become more prevalent in the lower ranks.  

Table~\ref{table:entropy_top_ten}  shows the kernels with the ten largest entropies and Table~\ref{table:entropy_bottom_ten} shows those with the ten smallest nonzero entropies.  We observe that the kernels with largest entropies are mostly of the form $(l_1l_2)$ and are almost perfectly balanced. 
The least balanced kernels tend to be more recognizable English or Spanish words and names, with one exclamation also appearing in the bottom ten.

\begin{table}[htbp!]
\begin{center}
\setlength\tabcolsep{5pt}
\begin{tabular}{|c|c|l|l|}
\hline
 & $H$ & Kernel & Example token\\
\hline
 1 & 0.99998 &  (kd) & kdkdkdkdkdkdkd\\
 2 & 0.99998 &  (ha) & hahahahahaha\\
 3 & 0.99997 &  [i][d] & iiiiiiiddddd\\
 4 & 0.99997 &  (ui) & uiuiuiuiuiui\\
 5 & 0.99997 &  (ml) & mlmlmlmlmlmlml\\
 6 & 0.99995 &  (js) & jsjsjsjsjsj\\
 7 & 0.99990 &  [e][t] & eeeeetttttt\\
 8 & 0.99988 &  (ox) & oxoxoxoxoxox\\
 9 & 0.99980 &  (xq) & xqxqxqxqxqxqxq\\
10 & 0.99971 &  (xa) & xaxaxaxaxaxa\\
\hline
\end{tabular}
\end{center}
\caption{Top 10 kernels by normalized entropy, $H$.}\label{table:entropy_top_ten}
\end{table}

\begin{table}[htbp!]
\begin{center}
\setlength\tabcolsep{5pt}
\begin{tabular}{|c|c|l|l|}
\hline
 & $H$ & Kernel & Example token\\
\hline
 1 & 0.01990 &  [b][o][b]ies & booooooobies\\
 2 & 0.02526 &  [d][o][d]e & doooooooode\\
 3 & 0.03143 &  infini[t][y] & infinityyyyy\\
 4 & 0.03342 &  che[l]se[a] & chelseaaaaaa\\
 5 & 0.03587 &  tay[l]o[r] & taylorrrrrr\\
 6 & 0.03803 &  f(re) & freeeeeeeeeeeee\\
 7 & 0.03930 &  [f]ai[r] & fairrrrrrrr\\
 8 & 0.05270 &  regr[e][s][e] & regreseeeeee\\
 9 & 0.05271 &  herm[a][n][a] & hermanaaaaaaaa\\
10 & 0.05323 &  sq[u][e] & squueeeeeeee\\
\hline
\end{tabular}
\end{center}
\caption{Bottom 10 kernels by normalized entropy, $H$.}\label{table:entropy_bottom_ten}
\end{table}

\subsection{Stretch}
\label{sec:stretch}

\begin{figure}[tp!]
\centering
\includegraphics[width=\columnwidth]{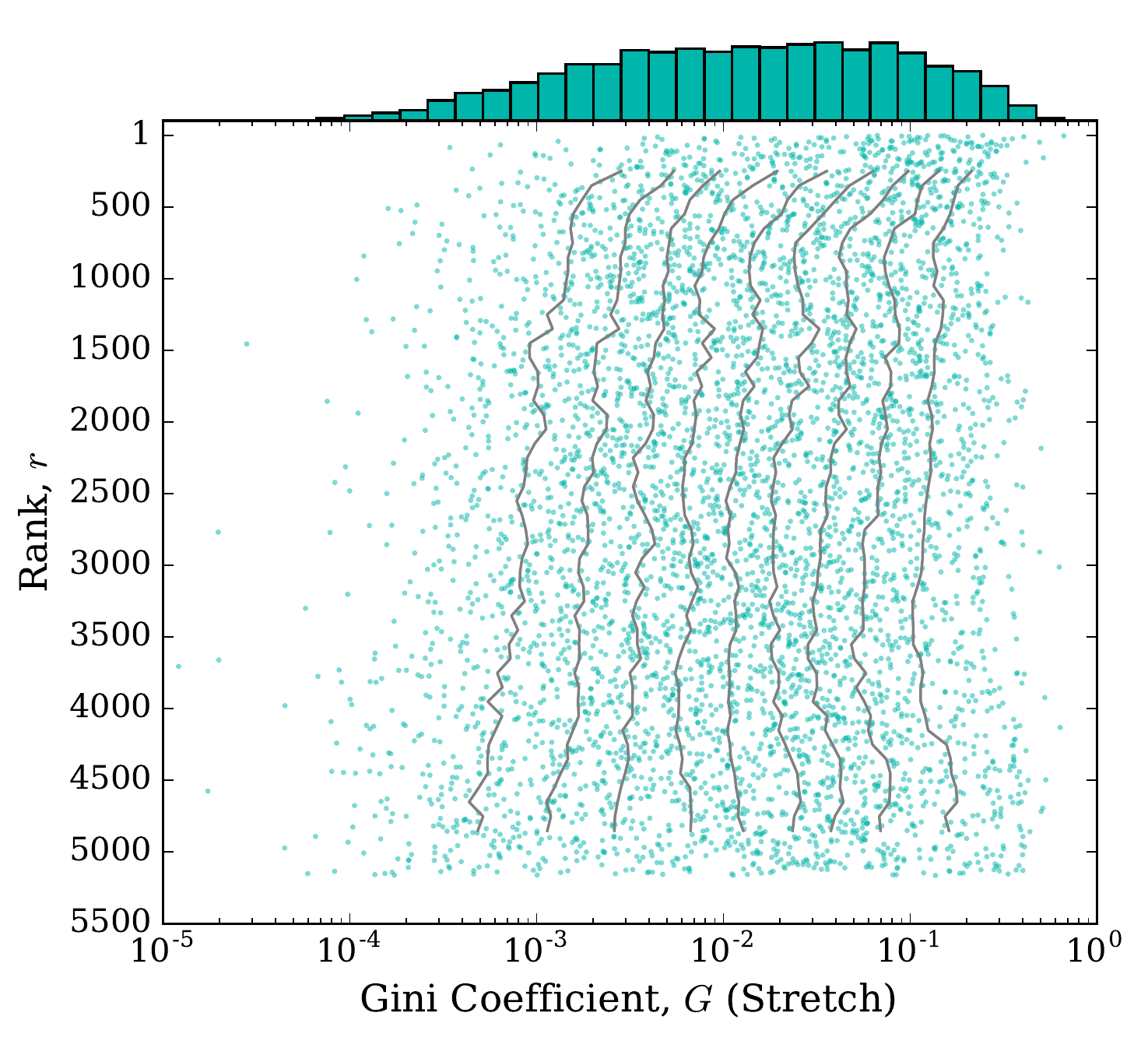}
\caption{Jellyfish plots for kernel stretch as measured by the Gini coefficient, $G$, of its token count distribution, where higher Gini coefficient denotes increased stretch.  The histogram is given at the top of the plot (with logarithmic width bins).  Kernels are plotted vertically by their rank, $r$, and horizontally (on a logarithmic scale) by their stretch.  The deciles $0.1, 0.2, \dots, 0.9$ are calculated for rolling bins of 500 kernels and are plotted as the `tentacles'.  }
\label{fig:gini_jellyfish}
\end{figure}

To measure overall stretchiness for a kernel we calculated the Gini coefficient, $G$, of the kernel's token length frequency distribution.  (For a comparison with another possible measure of stretch, see Appendix~\ref{appendix:ratio}.)  If the distribution has most of its weight on the short versions and not much on stretched out versions, then the Gini coefficient will be closer to 0.  If more tokens are long and the kernel is stretched longer more often, the Gini coefficient will be closer to 1.  Fig.~\ref{fig:gini_jellyfish} gives the jellyfish plot for the Gini coefficient for each kernel.  The horizontal axis has a logarithmic scale, and the histogram bins have logarithmic widths.  From this plot, we see that the distribution for stretch is quite stable across word ranks, except for perhaps a slight shift towards higher Gini coefficient (more stretchiness) for the highest ranked kernels.  

\begin{figure}[tp!]
\centering
\includegraphics[width=\columnwidth]{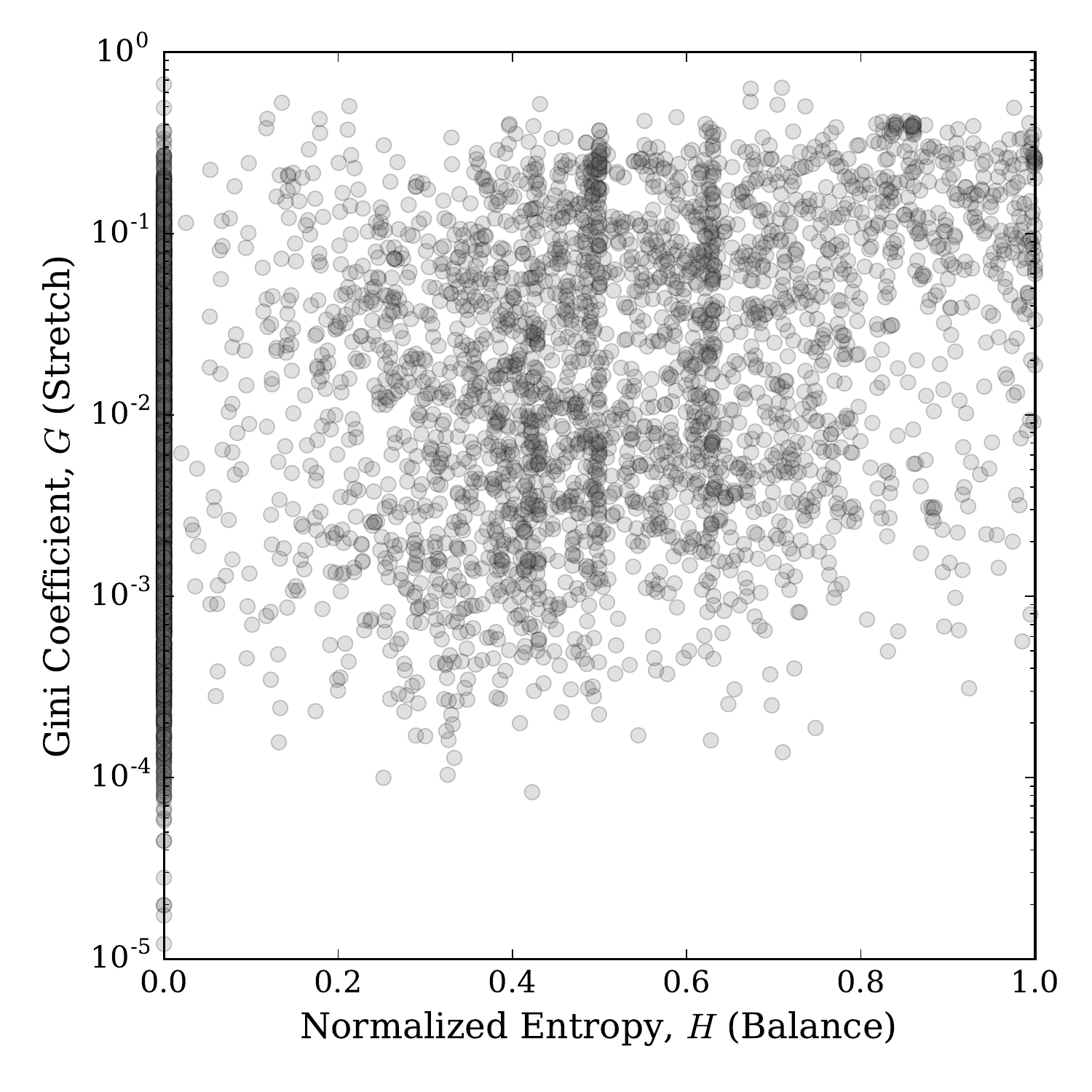}
\caption{Kernels plotted in Balance-Stretch parameter space.  Each kernel is plotted horizontally by the value of its balance parameter, given by normalized entropy, $H$, and vertically (on a logarithmic scale) by its stretch parameter, given by the Gini coefficient, $G$, of its token count distribution.  Larger entropy implies greater balance and larger Gini coefficient implies greater stretch. }
\label{fig:balance-and-stretch}
\end{figure}

\begin{table}[htbp!]
\begin{center}
\setlength\tabcolsep{5pt}
\begin{tabular}{|c|c|l|l|}
\hline
 & $G$ & Kernel & Example token\\
\hline
 1 & 0.66472 & [k] & kkkkkkkkkkkkkkk\\
 2 & 0.63580 & [w][v][w] & wwwwwwwwwwvwwww\\
 3 & 0.62843 & [m][n][m] & mmmmmmmmmmmmnm\\
 4 & 0.53241 & [o][c][o] & oooooooooco\\
 5 & 0.52577 & wa(ki) & wakikikikkkikikik\\
 6 & 0.51706 & (go)[l] & goooooooooool\\
 7 & 0.51273 & [m][w][m] & mmmmmwmmmmmmmmm\\
 8 & 0.50301 & galop[e]ir[a] & galopeeeeira\\
 9 & 0.50193 & [k][j][k] & kkkkkjjkkkkkkkkkk\\
10 & 0.49318 & [i][e][i] & iiiiiieeiiiiiii\\
\hline
\end{tabular}
\end{center}
\caption{Top 10 kernels by Gini coefficient, $G$.}\label{table:gini_top_ten}
\end{table}

\begin{table}[htbp!]
\begin{center}
\setlength\tabcolsep{5pt}
\begin{tabular}{|c|c|l|l|}
\hline
 & $G$ & Kernel & Example token\\
\hline
 1 & 0.00001 & am[p] & amppppppppp\\
 2 & 0.00002 & m[a]kes & maaaaaaaaakes\\
 3 & 0.00002 & fr[o]m & frooooooooooom\\
 4 & 0.00002 & watch[i]ng & watchiiiiiing\\
 5 & 0.00003 & w[i]th & wiiiiiiiith\\
 6 & 0.00004 & pla[y]ed & playyyyyyed\\
 7 & 0.00004 & s[i]nce & siiiiiiiince\\
 8 & 0.00006 & eve[r]y & everrrrrrrrrry\\
 9 & 0.00006 & manage[r] & managerrrrr\\
10 & 0.00007 & learnin[g] & learninggggg\\
\hline
\end{tabular}
\end{center}
\caption{Bottom 10 kernels by Gini coefficient, $G$.}\label{table:gini_bottom_ten}
\end{table}

Table~\ref{table:gini_top_ten} shows the top 10 kernels ranked by Gini coefficient and Table~\ref{table:gini_bottom_ten} shows the bottom 10.  The top kernel is [k], which represents laughter in Portuguese, similar to (ha) in English (and other languages).  Containing a single letter, [k] is easier to repeat many times, and does not have an unstretched version that is a common word.  We also see (go)[l] on the list, where `gol' is Spanish and Portuguese for `goal'.  Interestingly, (go)[l] has a much higher Gini coefficient ($G=0.5171$) than [g][o][a][l] does ($G = 0.1080$).  The kernels with lowest Gini coefficient all represent regular words and all allow just one letter to stretch, which does not get stretched much.  

\begin{figure*}[tp!]
\centering
\includegraphics[width=.98\textwidth]{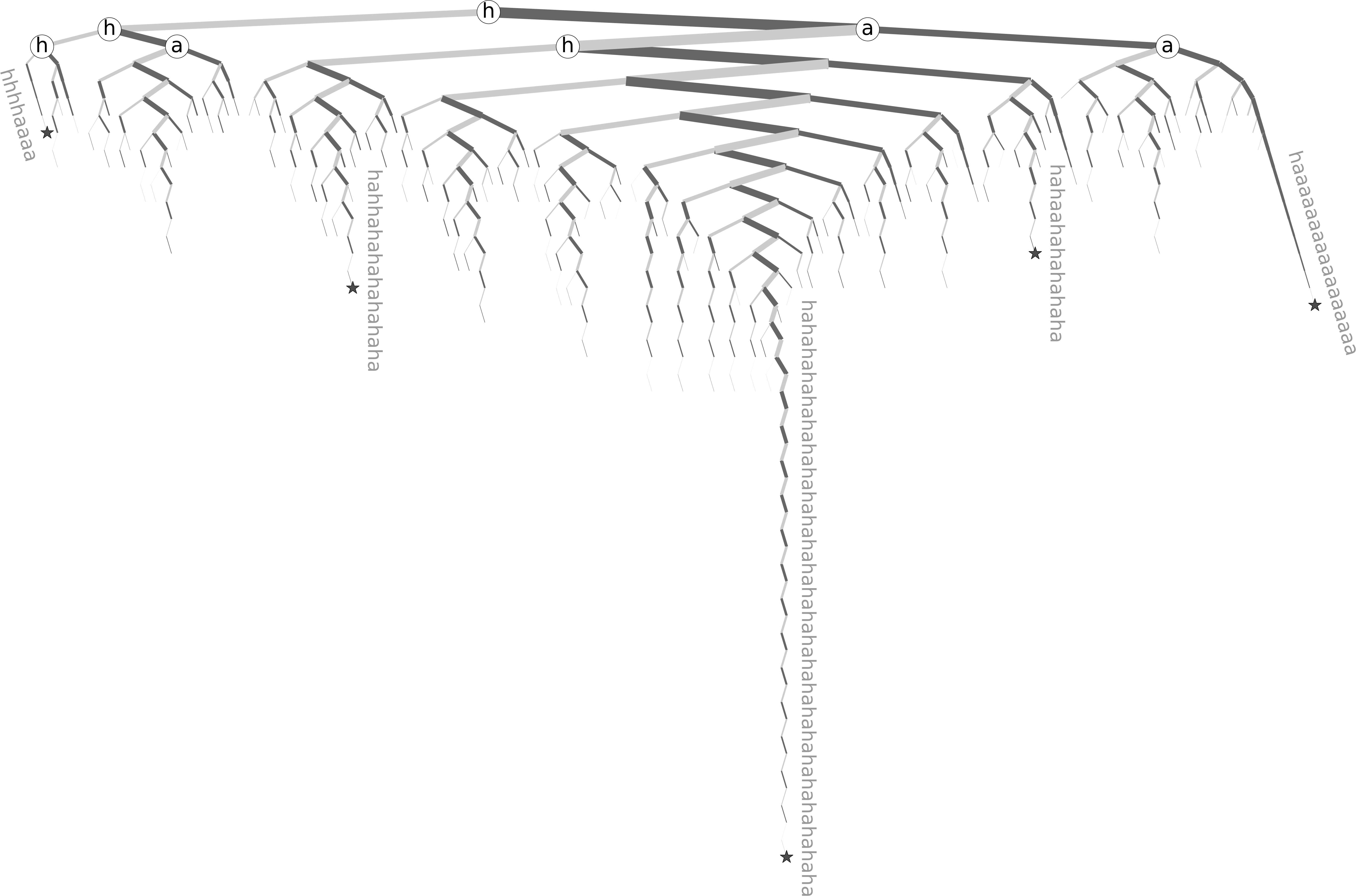}
\caption{\Wordtree for the kernel (ha).  The root node represents `h'.  From there, branching to the left (light gray edge) is equivalent to appending an `h'. Branching to the right (dark gray edge) is equivalent to appending an `a'.  The edge width is logarithmically related to the number of tokens that pass along that edge when spelled out.  A few example words are annotated, and their corresponding nodes are denoted with a star.  This tree was trimmed by only including words with a token count of at least 10,000.   The code used to create the figures for these \wordtrees is largely based on the algorithm presented by Wetherel and Shannon \cite{wetherell1979a}.  We note that Mill has written a more recent paper based largely on this earlier work specialized for Python \cite{mill2008a}, and an implementation for it as well \cite{mill2008b}, but they both contain algorithmic bugs (detailed in Appendix~\ref{appendix:algorithm_bugs}). }
\label{fig:ha_tree}
\end{figure*}

In Fig.~\ref{fig:balance-and-stretch}, we show a scatter plot of each kernel where the horizontal axis is given by the measure of balance of the kernel using normalized entropy,  and the vertical coordinate is given by the measure of stretch for the kernel using the Gini coefficient.  Thus, this plot positions each kernel in the two dimensional space of balance and stretch.  We see that the kernels spread out across this space and that these two dimensions capture two independent characteristics of each kernel.  

We do note that there are some structures visible in Fig.~\ref{fig:balance-and-stretch}.  There is some roughly vertical banding.  In particular, the vertical band at $H=0$ is from kernels that only allow one character to stretch and the vertical band near $H=1$ is from kernels where all characters are allowed to stretch and do so roughly equally, which especially occurs with kernels that are a single two letter element.  Fainter banding around $H\approx .43$, $H\approx .5$, and $H\approx .63$ can also be seen.  This largely comes from kernels of length 5, 4, and 3, respectively, that allow exactly two characters to stretch and those characters stretch roughly equally.  If the stretch was perfectly equal, then the normalized entropy in each respective case would be $H=1/\log_2(5) \approx .43$, $H=1/\log_2(4)=.5$, and $H=1/\log_2(3) \approx .63$.  

\begin{figure*}[tp!]
\begin{center}
\includegraphics[width=.92\textwidth]{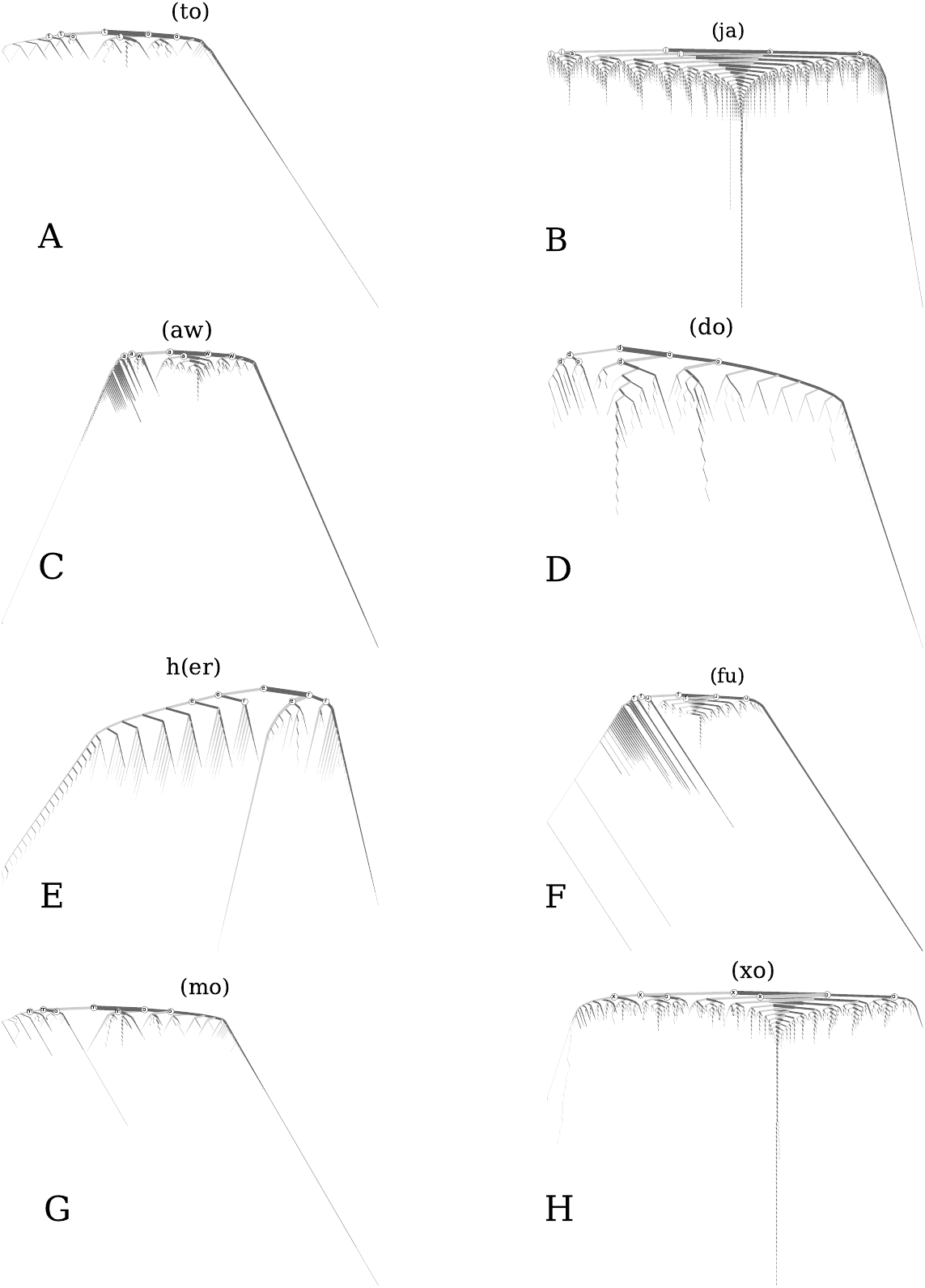}
\end{center}
\caption{A collection of example \wordtrees\unskip.  From left to right, top to bottom, trees for the kernels (to), (ja), (aw), (do), h(er), (fu), (mo), and (xo).  }
\label{fig:trees}
\end{figure*}

\subsection{\Wordtrees}
\label{sec:trees}

So far we have considered frequency distributions for kernels by token length, combining the token counts for all the different words of the same length matching the kernel.  However, different tokens of the same length may of course be different words---different stretched versions---of the same kernel.  For kernels that contain only single letter elements, these different versions may just have different amounts of the respective stretched letters, but all the letters are in the same order.  However, for kernels that have two letter elements, the letters can change order in myriad ways, and the possible number of different stretched versions of the same length becomes much larger and potentially more interesting.  

In order to further investigate these intricacies, we introduce `\wordtrees\unskip' to give us a visual method of studying the ways in which kernels with two letter elements are generally expanded.
Fig.~\ref{fig:ha_tree} gives the \wordtree for the kernel (ha).  
The root node is the first letter of the two letter element, which in this case is `h'.  Then, recursively, if the next letter in the word matches the first letter of the pair, it branches left, represented by a lighter gray edge, and if it matches the second letter of the pair then it branches right, represented by a darker gray edge.  This branching continues until the word is finished.  The first few nodes are highlighted with the letter corresponding to that point of the tree.  The edge weights are logarithmically related to the number of tokens flowing through them.  
In Fig.~\ref{fig:ha_tree}, a few nodes, denoted by stars, are annotated with the exact word to which they correspond.  The annotated nodes are all leaf nodes, but words can, and most do, stop at nodes that are not leaves.  
We also trimmed the tree by only including words that have a token count of at least 10,000. This threshold of pruning reveals the general pattern while avoiding making the \wordtree cluttered. 

The \wordtree for (ha) has a number of interesting properties.  Most notable among them is the self-similar, fractal-like structure.  The main branch line dropping down just right of center represents the perfect alternating sequence `hahahahaha\dots', as shown by the annotated example at the leaf of this line.  There are also many similar looking subtrees that branch off from this main branch that each have their own similar looking main branch.  These paths that follow the main branch, break off at one location, and then follow the main branch of a subtree represent words that are similar to the perfect alternating laugh, but either have one extra `h' (if the branch veers left) or one extra `a' (if the branch leads right).  For example, the middle left annotation shows that the fourth letter was an extra `h', and then the rest of the word retained an accurate alternating pattern. This word, `hahhahahahahahaha', appeared 13,894 times in our dataset.  

The tree also shows that `haaaaa\dots' is a strong pattern, as can be seen farthest right in the (ha) \wordtree\unskip.  The subtrees on the right show that users also start with the back and forth pattern for a stretch, and then finish the word with trailing `a's.   Many other patterns also appear in this tree, and additional patterns are occluded by our trimming of the tree, but likely most of these come from users trying to follow one of the patterns we have already highlighted and introducing mistypings. 

We made similar trees for every kernel that had a single occurrence of a two letter element, where the tree represents just the section of word that matches the two letter element.  These trees are trimmed by only including words that have a token count of at least the fourth root of the total token count for the stretched tokens.

Fig.~\ref{fig:trees} gives eight more examples of these \wordtrees\unskip.  The trees for (ja) and (xo) have many of the same characteristics as the tree for (ha), as do most of the trees for kernels that are a two letter element where tokens predominantly alternate letters back and forth.  For the tree for (xo), the pattern where the first letter of the two letter element is stretched, followed by the second letter being stretched, such as `xxxxxooooo', is more apparent.  This type of pattern is even more notable in the trees for (aw), and especially (fu).  The tree for (mo) has stretched versions for both `mom' and `moo'.  Similarly, the tree for h(er) shows stretched versions of both `her' and `here', where we see that both `e's and the `r' all get stretched.  In the tree for (to), the word `totoo' has a much larger token count then words stretched beyond that (noticeable by the fact that the edges leaving that node are much smaller than the edge coming in).  The word `totoo' is Tagalog for `true'.  Finally, every example tree here does show the back and forth pattern to at least some extent.  All of the trees created are available for viewing in Online Appendix D at \onlineappendixurl.  

\begin{figure*}[tp!]
\centering
\includegraphics[width=\textwidth]{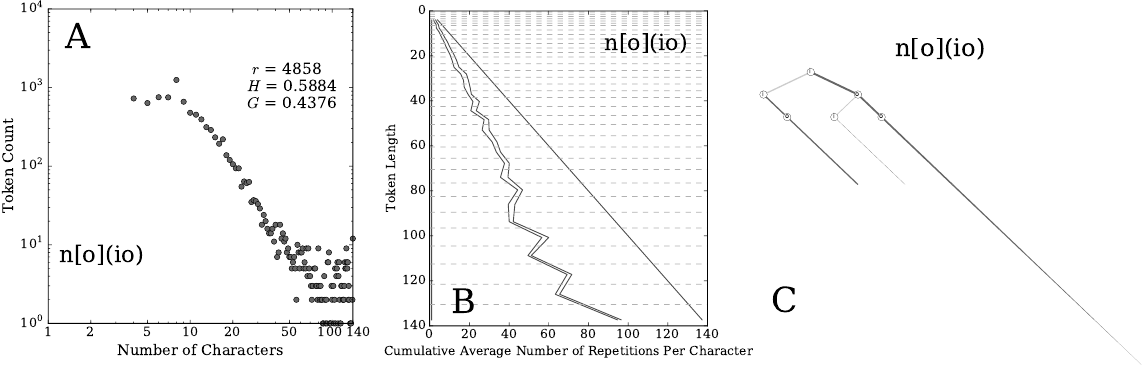}
\caption{(A) Token count distribution, (B) balance plot, and (C) \wordtree for the kernel n[o](io).  In general, these types of plots offer diagnostic help when studying mistypings.  In this case, they provide evidence towards the conclusion that the words that match this kernel were likely meant to be stretched versions of the word `no' with a few mistaken `i's included. Note that `i' is next to `o' on a standard QWERTY keyboard. }
\label{fig:noio}
\end{figure*}

\subsection{Mistypings}
\label{sec:mistypings}

Mistypings appear often in tweets and we see evidence of them in stretched words.  For example, the kernel n[o](io) is likely a result of mistypings of n[o].  On at least some platforms, holding down the key for a letter does not make that letter repeat, so one must repeatedly press the same key.  For the standard QWERTY keyboard layout, the letter `i' is next to the letter `o', so it would be easy to accidentally press the letter `i' occasionally instead of `o' when trying to repeat it many times, especially on the small keyboards accompanying mobile phones.  This sort of thing could lead to a kernel like n[o](io) when users try to stretch the word `no'.  Similarly, the letters `a' and `s' are next to each other on a QWERTY keyboard, so a kernel like (ha)s(ha)(sh)(ah) likely comes from mistypings of the much simpler kernel (ha). 

However, it is not always clearly apparent if a kernel is from mistypings or on purpose, or perhaps comes as a result of both.  For example, the letter `b' is close to the letter `h', so the kernel (ha)b(ah) could come from mistypings of (ha). But, this form could also be intentional, and meant to represent a different kind of laughter. For example, (ba)(ha) is a highly ranked kernel (rank 211) representing a comedically sinister kind of laughter.  Similarly, (ja) is a core component of laughter in Spanish, but `j' is next to `h' on the QWERTY keyboard, so it is not apparent if a kernel like (ha)j(ah)(ja)(ha) comes from mistypings or from switching back and forth between English and Spanish as the word stretches.  

Our methodology may enable further study of mistypings.  For example, Fig.~\ref{fig:noio} shows the distribution, balance plot, and \wordtree for the kernel n[o](io).  The distribution shows that it is not a strong kernel, with the lower rank of 4,858, compared to a rank of 8 for (no).  The balance plot shows that the letter `i' is not stretched much, and the \wordtree shows that the word is mostly just a repetition of `o's.  
On the whole, the evidence suggests that the kernel n[o](io) is mainly a result of mistypings.

\begin{figure}[htp!]
\centering
\includegraphics[width=0.80\columnwidth]{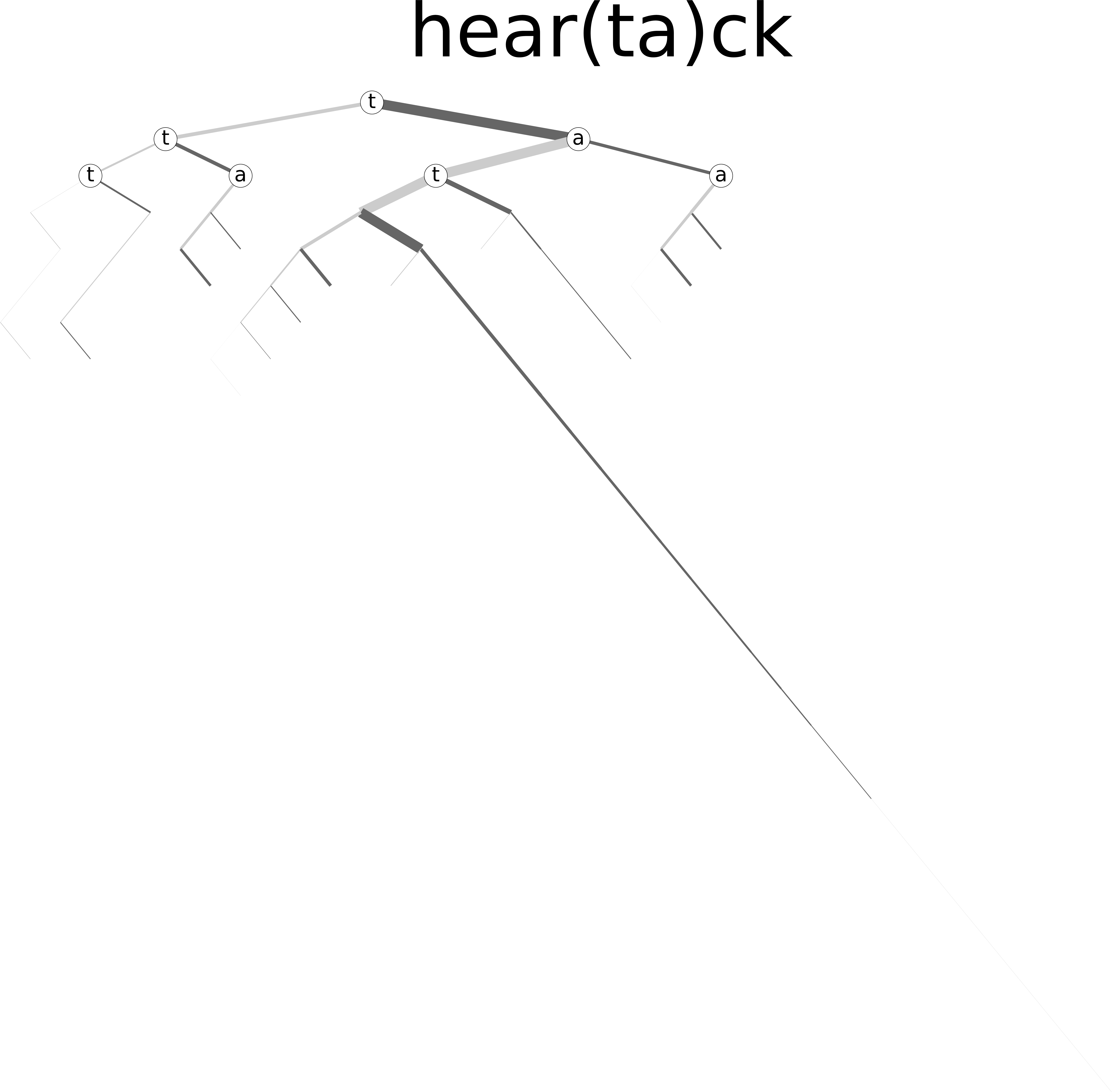}
\caption{\Wordtree for the kernel hear(ta)ck.  From this tree, we can see the relative number of times the word `heartatack' is written rather than `heartattack', indicating a common misspelling. }
\label{fig:heartattack_tree}
\end{figure}

These tools can also be used to help study what are likely misspellings, rather than mistypings.  For example, Fig.~\ref{fig:heartattack_tree} shows the \wordtree for the kernel hear(ta)ck (which does not actually fall within our rank cutoff, as described in Sec.~\ref{sec:data}, but provides a good example).  The word `attack' has two `t's.  Thus, the word `heartattack' (if written as one word; usually it is two) should, under normal spelling, have a double `t' after the second `a'.  From Fig.~\ref{fig:heartattack_tree} we can see from the weights of the branches that it is often written as `heartatack', with a single `t' instead of the double `t'.  

\section{Concluding remarks}
\label{sec:concludingremarks}

In this paper, we have studied stretched words, which are often used in spoken language. Until the advent of social media, stretched words were not prevalent in written language and largely absent from dictionaries.  The area of stretchable language is rich, and we have discovered that these words span at least the two dimensional parameter space of balance and stretch.  

The tools we have developed not only help uncover the hidden dynamics of stretchable words, but can be further applied to study phenomena such as mistypings and misspellings, and possibly more.  
Online dictionaries, such as the Wiktionary \cite{wiktionary}, could use our kernels as a general entry for each type of stretchable word, and include the balance and stretch parameters as part of their structured word information, as they do, for example, with part of speech.  
Natural language processing software and toolkits could use the techniques we developed to help with processing stretched words, e.g., in their approach to stemming.
Similarly, spell checking software may be able to use our methods to help prevent marking stretched words as misspellings.
Our procedures could also be used to help prevent typosquatting \cite{wiki:typosquatting}.  
Twitter could use our methods to help improve their spam filter, looking for slight variations of tweets.  
Also, \wordtrees could more generally be used to analyze the construction of any sequence, such as genome sequences.

However, much more could be done. 
We have restricted our study to words containing only Latin letters. Future work could extend this to include all characters, including punctuation and emojis.  We also limited the way we constructed kernels, focusing only on one and two letter elements.  This can be expanded to three letter elements and possibly beyond to capture the characteristics of words like `omnomnomnom'.  Furthermore, our methodology for creating kernels leads to situations where, for example, we have both (ha)g(ah) and (ha)(ga)(ha) as kernels.  Expanding to three letter elements and beyond in the future could collapse these forms, and related kernels, into a kernel like (hag).  

Along with more advanced kernels, similar but more advanced \wordtrees could be developed.  We only created \wordtrees for kernels with a single two letter element. Future work could explore kernels with more than two letter elements.  They could also be created for every kernel, where the branching of even the single letter elements is shown, where one branch would signify the repetition of that letter and the other branch would signify moving onto the next letter of the kernel.  Furthermore, to go with three letter elements, ternary trees could be developed.  Among other things, this would reveal mistypings like (ha)(hs), for example, if this became a kernel with a three letter element like (has), and we assume that the `s' is mostly a mistyping of the letter `a' in the kernel (ha). This situation should be discernible from the case where the word `has' is stretched.  

Another interesting phenomenon to look at is the distinction between phonetic and visual stretching.  When verbally stretching a word, only certain sounds can be stretched out, whereas when typing a word, any letter can be repeated.  For example, compare `gooooaaaaal' with `ggggggooooooooaaaaaalllll'.  Both stretched words can be typed, but only the first can be said because of the plosive `g'.  Relatedly, looking at what parts of words, such as the end of words, or which letters get stretched more could be interesting.  

Finally, our methodology could be used to explore linguistic and behavioral responses to changes in Twitter's protocol (e.g., character length restrictions) and platform (e.g., mobile vs.~laptop). For example, what are the effects of auto-correct, auto-complete, and spell check technologies?  And what linguistic changes result from platform restrictions such as when a single key cannot be held down anymore to repeat a character? Also, we only considered tweets before the shift from the 140 to 280 character limit on Twitter.  Some initial work indicates that the doubling of tweet length has removed the edge effect that the character limit creates \cite{gligoric2018a_conference}.  Further work could study how this change has affected stretchable words, and in particular, the tail of their distributions. \\

\acknowledgments
CMD and PSD were supported by NSF Grant No.~IIS-1447634, and TJG, CMD, and PSD were supported by a gift from MassMutual.

\bibliographystyle{unsrt}

\clearpage

\onecolumngrid
\appendix

\setcounter{table}{0}
\renewcommand{\thetable}{\Alph{section}\arabic{table}}
\setcounter{figure}{0}
\renewcommand{\thefigure}{\Alph{section}\arabic{figure}}

\newpage
\section{Alternate Balance Measure}
\label{appendix:entropy_alternate}
\twocolumngrid

\begin{figure}[bp!]
\centering
\includegraphics[width=\columnwidth]{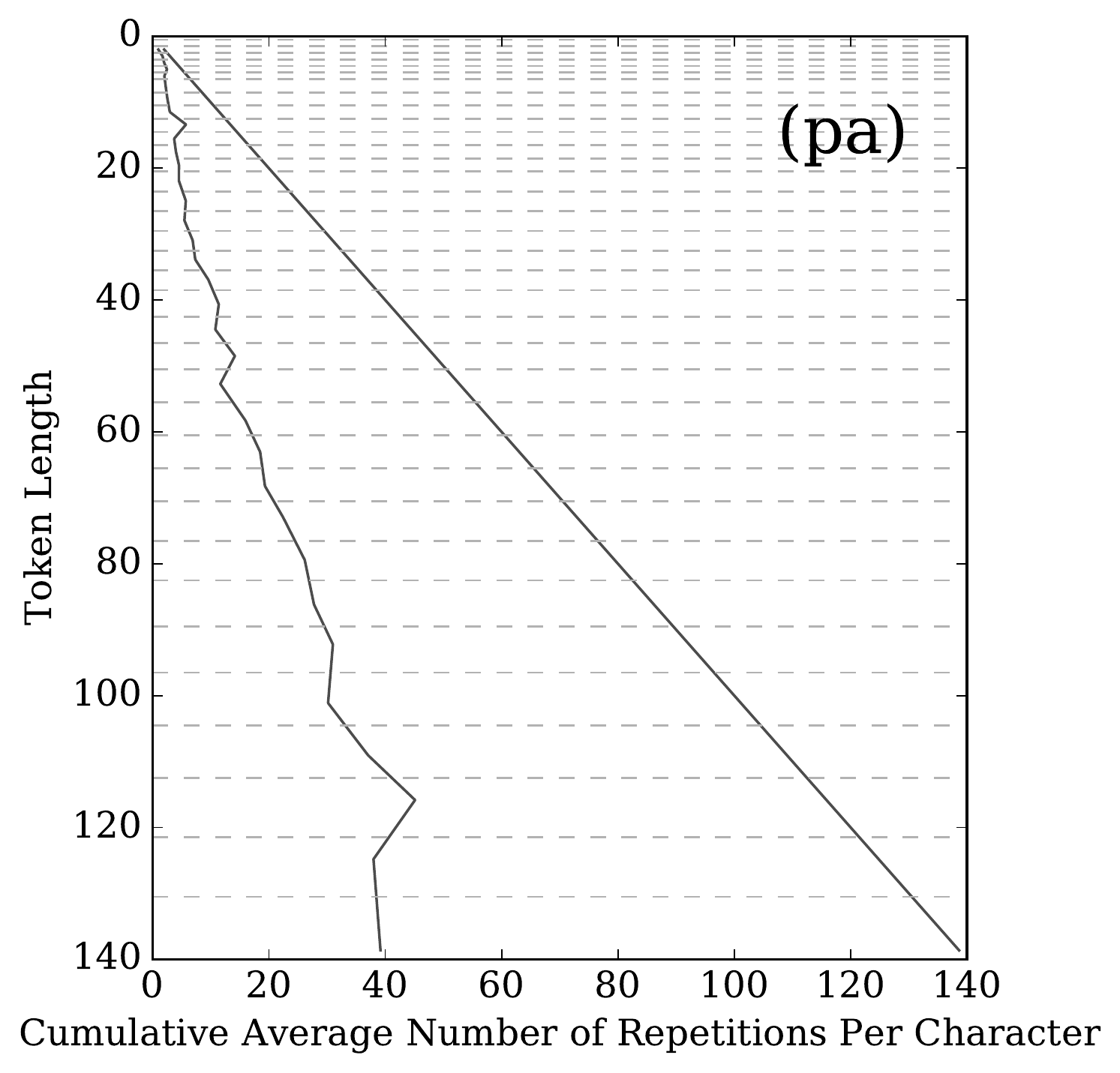}
\caption{Balance plot for the kernel (pa).  See the Fig.~\ref{fig:ha-balance} caption for plot details.  Even though $H_\textnormal{alt}=1.00000$ for (pa), this plot clearly shows perfect balance is not sustained as tokens increase in length.  }
\label{fig:pa-balance}
\end{figure}

\begin{figure}[tp!]
\centering
\includegraphics[width=\columnwidth]{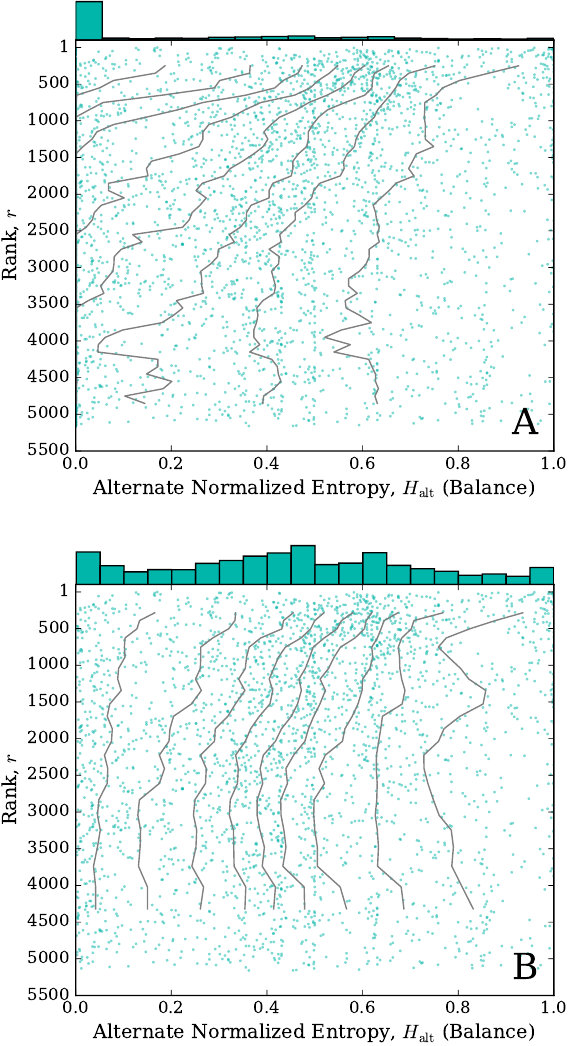}
\caption{Jellyfish plots for kernel balance based on an alternate entropy measure for (A) all kernels, and (B) excluding kernels with entropy exactly 0.  Corresponding histograms are given at the top of each plot.  Kernels are plotted vertically by their rank, $r$, and horizontally by their balance as given by an alternate normalized entropy, $H_{\textnormal{alt}}$, where larger entropy denotes increased balance.  The deciles $0.1, 0.2, \dots, 0.9$ are calculated for rolling bins of 500 kernels and are plotted as the `tentacles'. }
\label{fig:balance_jellyfish_alt}
\end{figure}

As a comparison to our normalized entropy measure for balance discussed in Sec.~\ref{sec:balance}, we also compute an alternate normalized entropy measure, $H_{\textnormal{alt}}$, that measures balance from a different view.  

To compute $H_{\textnormal{alt}}$, we first calculate the overall average stretch for each character as before, but now do so across all tokens at once.  Then, we subtract one from each of these values and normalize them so they sum to 1 and can be thought of like probabilities.  We then compute the normalized entropy, $H_{\textnormal{alt}}$, of these values as a measure of overall balance.  $H_\textnormal{alt}$ is similar to $H$ in that if each character stretches the same on average, the normalized entropy is 1, and if only one character in the kernel stretches, the normalized entropy is 0.  Again, higher entropy corresponds with more balanced words.  

The difference is the view, and what is meant by `on average'.  For $H_\textnormal{alt}$, each token is weighted equally when calculating balance.  Thus, this measure corresponds to the view of if one randomly samples tokens and looks at how balanced they are on average.

By contrast, for $H$, as calculated in Sec.~\ref{sec:balance}, tokens are grouped by length, and then each group gets an equal weight regardless of the group size.  This view looks at how well balance is sustained across lengths, and corresponds to sampling tokens by first randomly picking a length, and then randomly picking a token from all tokens of that length, and then looking at how balanced the sampled tokens are on average.  

For example, for the kernel (pa), $H_\textnormal{alt} = 1.00000$, signifying nearly perfect balance.  However, looking at the balance plot for (pa) in Fig.~\ref{fig:pa-balance}, we see that perfect balance is not sustained across lengths.  Because most of the tokens are short, and short stretched versions of (pa) are well balanced, all of the weight is on the well balanced short ones when randomly picking tokens.  However, as people create longer stretched versions of (pa), they tend to use more `a's than `p's, and near perfect balance is not maintained. 
This is better captured by the measure $H = 0.80982$.

As our main measure of balance, we chose the view better representing how well balanced tokens are as they are stretched, equally weighing lengths.  This does have the limitation that groups of tokens with different lengths have different sizes, and some of them may contain a single token, possibly increasing the variance of the measure.  It is possible this could be improved in the future by only including lengths that have a certain number of examples, or possibly creating larger bins of lengths for the longer tokens like we do in the balance plots.   

\begin{table}[tp!]
\begin{center}
\setlength\tabcolsep{5pt}
\begin{tabular}{|c|c|l|l|}
\hline
 & $H_{\textnormal{alt}}$ & Kernel & Example token\\
\hline
 1 & 1.00000 &  (ba) & baaaaaaaaaaa\\
 2 & 1.00000 &  (pa) & ppppppppppppa\\
 3 & 1.00000 &  (uo) & uouuuuuuuuuuuuu\\
 4 & 0.99998 &  (pr) & prrrrrrrrrrr\\
 5 & 0.99998 &  (du) & duduudduududududuuu\\
 6 & 0.99995 &  (xa) & xaxaxaxaxxa\\
 7 & 0.99995 &  (ai) & aaaaaaaaaaaaaaaai\\
 8 & 0.99993 &  (he) & hehehheheheh\\
 9 & 0.99986 &  (bi) & biiiiiiiiii\\
10 & 0.99985 &  (wq) & wqwqwqwqwqw\\
\hline
\end{tabular}
\end{center}
\caption{Top 10 kernels by an alternate normalized entropy, $H_{\textnormal{alt}}$.}\label{table:entropy_top_ten_alt}
\end{table}

\begin{table}[tp!]
\begin{center}
\setlength\tabcolsep{5pt}
\begin{tabular}{|c|c|l|l|}
\hline
 & $H_{\textnormal{alt}}$ & Kernel & Example token\\
\hline
 1 & 0.00115 &  [t][e][t]h & teeeeeeeeeth\\
 2 & 0.00119 &  f[e]l[i]ng & feeeeeeling\\
 3 & 0.00170 &  c[a][l]ing & calllllling\\
 4 & 0.00196 &  a[c]ep[t] & accepttttttt\\
 5 & 0.00197 &  fa[l][i]ng & falllllling\\
 6 & 0.00217 &  hi[l]ar[y] & hilllllaryy\\
 7 & 0.00227 &  m[i][s][i]ng & missssssssssing\\
 8 & 0.00271 &  ba[n]e[d] & baneddddddddd\\
 9 & 0.00277 &  t[h][r][e] & threeeeeeeee\\
10 & 0.00302 &  th(er) & therrrreeeee\\
\hline
\end{tabular}
\end{center}
\caption{Bottom 10 (nonzero) kernels by an alternate normalized entropy, $H_{\textnormal{alt}}$.}\label{table:entropy_bottom_ten_alt}
\end{table}

\begin{figure}[bp!]
\centering
\includegraphics[width=\columnwidth]{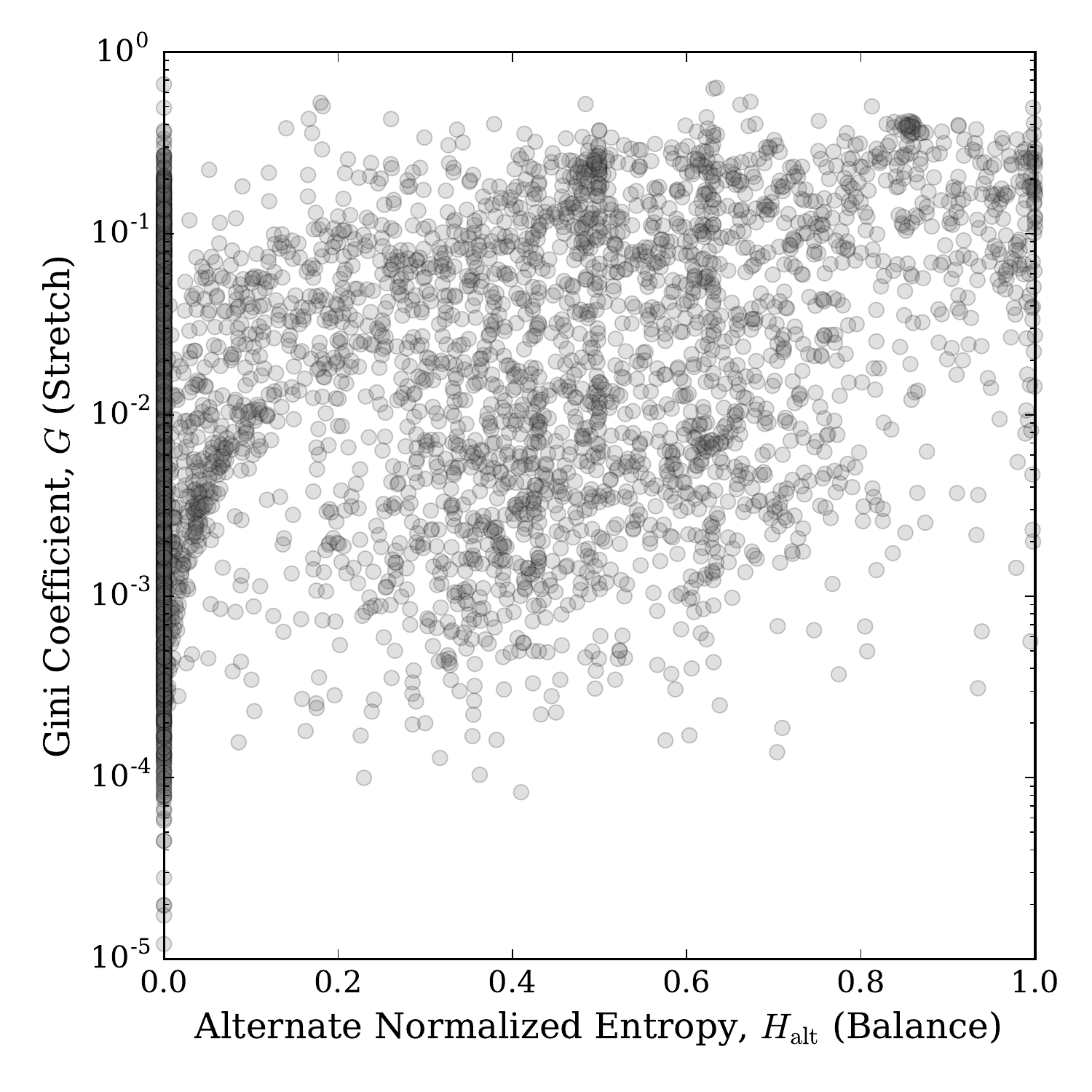}
\caption{Kernels plotted in Balance-Stretch parameter space using an alternate measure of normalized entropy for balance.  Each kernel is plotted horizontally by the value of its balance parameter, given by an alternate normalized entropy, $H_{\textnormal{alt}}$, and vertically (on a logarithmic scale) by its stretch parameter, given by the Gini coefficient, $G$, of its token count distribution.  Larger entropy implies greater balance and larger Gini coefficient implies greater stretch. }
\label{fig:balance-and-stretch_alt}
\end{figure}

We include the same plots and tables for $H_\textnormal{alt}$ as we did with $H$, and many of the observations are similar.
Fig.~\ref{fig:balance_jellyfish_alt} shows the two jellyfish plots for $H_\textnormal{alt}$.  Similar to before, Fig.~\ref{fig:balance_jellyfish_alt}A is the version containing all words and for Fig.~\ref{fig:balance_jellyfish_alt}B we remove the words that have a value of 0 for entropy.  The top of the plots in Fig.~\ref{fig:balance_jellyfish_alt} shows the frequency histograms in each case.  
As before, after removing kernels with an entropy of 0, we see a small left-shift in the highest ranked kernels, and then the distribution largely stabilizes.  Again, the highest ranked kernels tend to be more equally balanced, and kernels only stretching a single character tend to be lower ranked.  

Table~\ref{table:entropy_top_ten_alt}  shows the kernels with the ten largest entropies and Table~\ref{table:entropy_bottom_ten_alt} shows those with the ten smallest nonzero entropies as measured in this alternate way.  We observe that the kernels with largest entropies are all of the form $(l_1l_2)$ and are almost perfectly balanced given the view of equally weighing all tokens.  The kernels with lowest entropies all expand to regular words that when spelled in the standard way contain a letter that is repeated, plus these kernels allow other letters to stretch.  

Finally, Fig.~\ref{fig:balance-and-stretch_alt} shows the scatter plot of each kernel where the horizontal axis is given by this alternate measure of balance, $H_\textnormal{alt}$,  and the vertical coordinate is again given by the measure of stretch for the kernel using the Gini coefficient, $G$.  We again see that the kernels span the two dimensional space. 

We still get the same kind of rough vertical banding that we saw in Fig.~\ref{fig:balance-and-stretch} for the same reason, but we also see a curved dense band at lower entropy values, which seems to mostly contain kernels whose base word is spelled with a double letter, like `summer' (with kernel [s][u][m][e][r]).  

\onecolumngrid

\clearpage

\setcounter{table}{0}
\setcounter{figure}{0}

\clearpage
\newpage
\section{Stretch Ratio}
\label{appendix:ratio}
\twocolumngrid

\begin{figure}[bp!]
\centering
\includegraphics[width=\columnwidth]{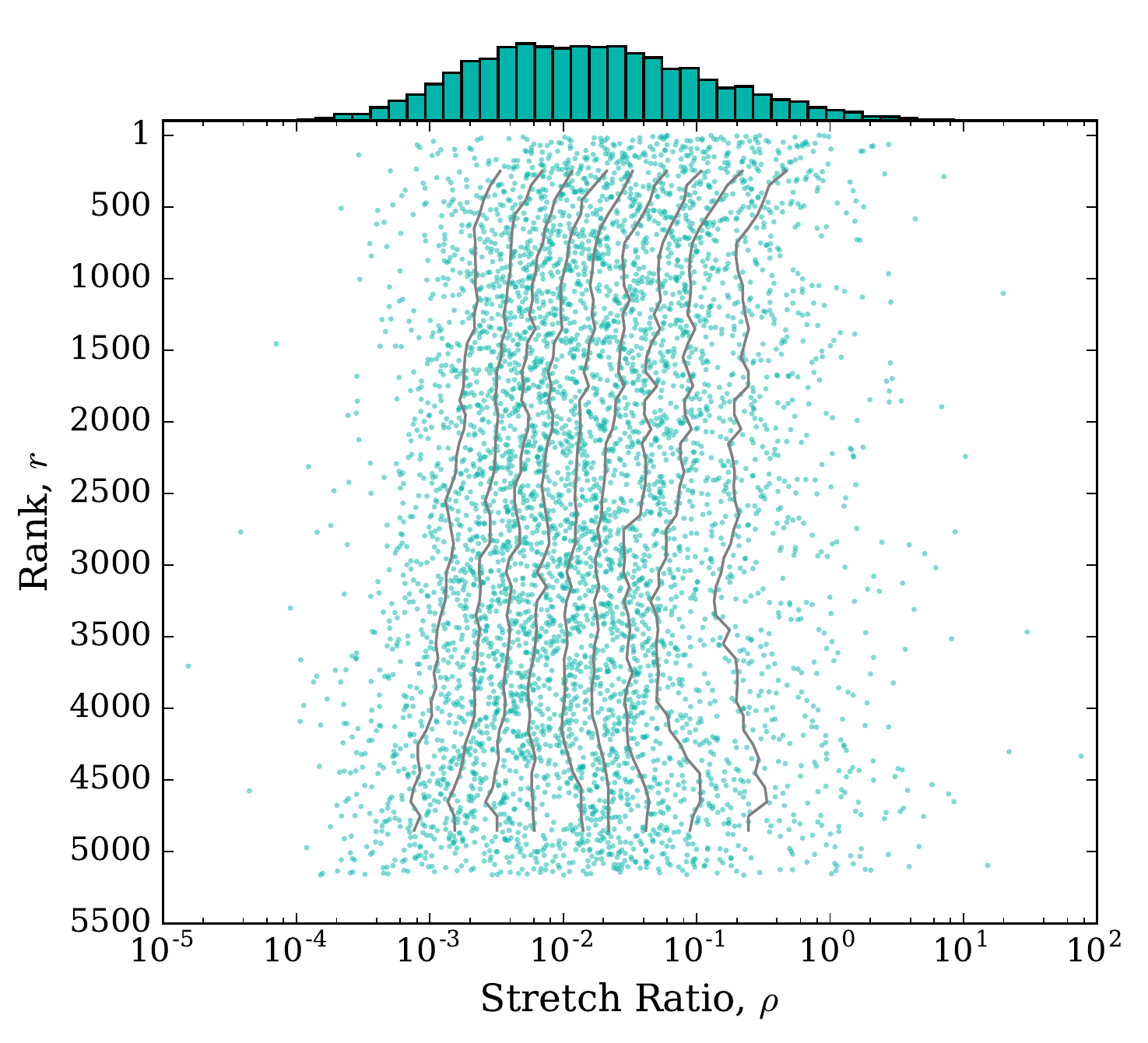}
\caption{Jellyfish plots for kernel stretch ratio, $\rho$, as given by the ratio of the sum of the kernel's stretched tokens to the sum or its unstretched tokens.  The histogram is given at the top of the plot (with logarithmic width bins).  Kernels are plotted vertically by their rank and horizontally (on a logarithmic scale) by their stretch ratio.  The deciles $0.1, 0.2, \dots, 0.9$ are calculated for rolling bins of 500 kernels and are plotted as the `tentacles'.  }
\label{fig:ratio_jellyfish}
\end{figure}

\begin{figure}[tp!]
\centering
\includegraphics[width=\columnwidth]{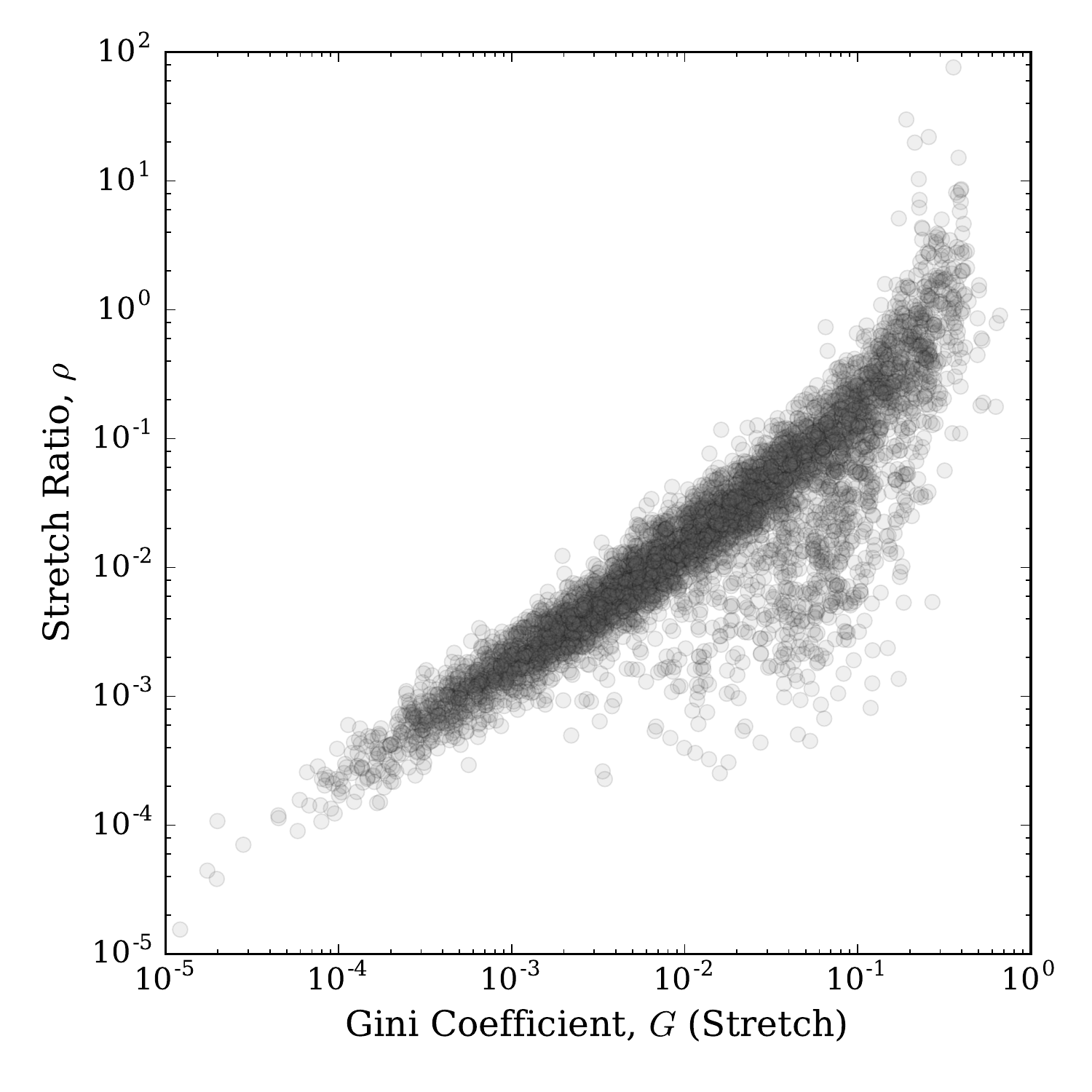}
\caption{Scatter plot of measures of stretch for each kernel.  For each kernel, the horizontal axis gives its stretch as measured by the Gini coefficient, $G$, of its token count distribution and the vertical axis gives its stretch ratio, $\rho$.  Both axes have a logarithmic scale. }
\label{fig:gini_vs_ratio}
\end{figure}

For each kernel, we also measure a `stretch ratio', $\rho$.  This is simply the ratio of the total number of stretched tokens, $n_\textnormal{s}$, to the total number of unstretched tokens, $n_\textnormal{u}$, for that kernel. 
That is,
\begin{equation}
\rho = \frac{n_\textnormal{s}}{n_\textnormal{u}}.
\end{equation}
Fig.~\ref{fig:ratio_jellyfish} gives the jellyfish plot for the stretch ratio.  Like Fig.~\ref{fig:gini_jellyfish}, the horizontal axis has a logarithmic scale and the histogram bins have logarithmic widths.  The stretch ratio distribution stays fairly stable across ranks, except for the highest ranked kernels, which tend to have a larger ratio.  

This stretch ratio can be thought of as a simple measure for the stretchiness of a kernel, with a larger ratio representing stretchier words.  As stretched versions of the word are used more, the numerator increases and the ratio value increases.  Conversely, as unstretched versions of the kernel are used more, the denominator increases, and the ratio value decreases.  
However, this simpler measure uses less information from the full distribution than a measure like the Gini coefficient does, so we would expect some differences between the two.
Indeed, Fig.~\ref{fig:gini_vs_ratio} shows that there are some kernels for which the two measures seem to disagree.  Yet, Fig.~\ref{fig:gini_vs_ratio} shows that the stretch ratio and Gini coefficient are quite well correlated, with Pearson correlation coefficient 0.89 ($p < 10^{-100}$), so there is not much gained by including both.  We choose to use the Gini coefficient as our main measure of stretchiness both because of its wide usage and because of the fact that it uses more information from the full distribution than the simpler stretch ratio.  

\begin{table}[tp!]
\begin{center}
\setlength\tabcolsep{5pt}
\begin{tabular}{|c|c|l|l|}
\hline
 & $\rho$ & Kernel & Example token\\
\hline
 1 & 76.04717 &  s[o][c][o][r][o][k] & socorrokkkkkk\\
 2 & 29.94863 &  mou(ha) & mouhahahaha\\
 3 & 21.93369 &  p[f](ha) & pffhahahaha\\
 4 & 19.82821 &  bu(ha) & buhahahahaha\\
 5 & 15.15702 &  (ha)j(ah)(ja)(ha) & hahahahajahajaha\\
 6 & 10.32701 &  pu(ha) & puhahahahaa\\
 7 & 8.63055 &  (ha)(ba)(ha) & habahahhaha\\
 8 & 8.47429 &  (ha)b(ha) & hahahhahabha\\
 9 & 8.13269 &  (ah)j(ah) & ahahahjahah\\
10 & 7.72953 &  a[e]h[o] & aehooooooooooooo\\
\hline
\end{tabular}
\end{center}
\caption{Top 10 kernels by stretch ratio, $\rho$.}\label{table:ratio_top_ten}
\end{table}

\begin{table}[tp!]
\begin{center}
\setlength\tabcolsep{5pt}
\begin{tabular}{|c|c|l|l|}
\hline
 & $\rho$ & Kernel & Example token\\
\hline
 1 & 0.00002 &  am[p] & amppppppppp\\
 2 & 0.00004 &  fr[o]m & froooooooom\\
 3 & 0.00004 &  m[a]kes & maaaaaaakes\\
 4 & 0.00007 &  w[i]th & wiiiiiiiiiiiiith\\
 5 & 0.00009 &  eve[r]y & everrrrrrrrrry\\
 6 & 0.00011 &  p[r]a & prrrrrrrrrrra\\
 7 & 0.00011 &  watch[i]ng & watchiiiing\\
 8 & 0.00011 &  s[i]nce & siiiiiiiince\\
 9 & 0.00012 &  pla[y]ed & playyyyyyyed\\
10 & 0.00012 &  vi[a] & viaaaaaaaaaaaaaaa\\
\hline
\end{tabular}
\end{center}
\caption{Bottom 10 kernels by stretch ratio, $\rho$.}\label{table:ratio_bottom_ten}
\end{table}

Table~\ref{table:ratio_top_ten} shows the top 10 kernels by stretch ratio and Table~\ref{table:ratio_bottom_ten} gives the bottom 10.  The correlation between stretch ratio and Gini coefficient, at least for the least stretchy kernels, can be seen further when comparing this to Table~\ref{table:gini_bottom_ten}.  Many of the kernels that show up as the least stretchy words (lowest Gini coefficients) also show up here in the list of kernels with smallest stretch ratio.

\onecolumngrid

\setcounter{table}{0}
\setcounter{figure}{0}

\clearpage
\newpage
\section{``Drawing presentable trees'' algorithmic bugs}
\label{appendix:algorithm_bugs}
\twocolumngrid

Wetherel and Shannon presented an algorithm for drawing large trees in a nice way in their paper ``Tidy drawing of trees'' \cite{wetherell1979a}.  
The article ``Drawing presentable trees'' \cite{mill2008a} by Mill and related code \cite{mill2008b}, based largely on the earlier work of Wetherel and Shannon, provide a version of the algorithm written in the Python syntax, but both the article and the code contain algorithmic bugs.  In the following, we present the bugs we found.

We will discuss Listing 5 in Mill's paper \cite{mill2008a}, as that is the version that most closely resembles Algorithm 3 of Wetherel's and Shannon's paper \cite{wetherell1979a}, which is what our code to create the \wordtrees is based off of.  

In Listing 5, the definition of \texttt{setup} contains the code:
\begin{verbatim}
elif len(tree.children) == 1:
    place = tree.children[0].x - 1
\end{verbatim}

This needs to be split into a left case and a right case.  If the only child node is a left child, then the parent should be placed to the right by one, and if the only child node is a right child, then the parent should be placed to the left by one.  The \texttt{DrawTree} class needs a way to tell if a node has a left or right child.  Let us assume the class \texttt{DrawTree} has an attribute \texttt{left} properly implemented that is set to \texttt{True} iff the node has a left child.  Then the code should be something more like the following:
\begin{verbatim}
elif len(tree.children) == 1:
    if tree.left:
        place = tree.children[0].x + 1
    else:
        place = tree.children[0].x - 1
\end{verbatim}

Compare the above fix to the corresponding code in the \texttt{right\_visit} case in the first \texttt{while} loop in Algorithm 3 in ``Tidy drawing of trees'' \cite{wetherell1979a}:
\begin{alltt}
elseif current\(\uparrow\).left_son = nil
    then place := current\(\uparrow\).right_son\(\uparrow\).x - 1;
elseif current\(\uparrow\).right_son = nil
    then place := current\(\uparrow\).left_son\(\uparrow\).x + 1; 
\end{alltt}

Later in Listing 5 in the definition of \texttt{setup} is the following line:
\begin{verbatim}
nexts[depth] += 2
\end{verbatim}

However, we want the next available spot, recorded in \texttt{nexts}, to be two spots to the right of the current placement, and the current placement is sometimes different from the current next available spot.  Thus the line should look something like the following:
\begin{verbatim}
nexts[depth] = tree.x + 2
\end{verbatim}

Again, compare this to the corresponding code found near the end of the \texttt{right\_visit} case of the first \texttt{while} loop  of Algorithm 3 in ``Tidy drawing of trees'':
\begin{alltt}
next_pos[h] := current\(\uparrow\).x + 2;\\
\end{alltt}

The final bug in Listing 5 is not an algorithmic bug, but merely a typo.  In the definition of \texttt{addmods} is the line of code:
\begin{verbatim}
modsum += tree.offset
\end{verbatim}

However, \texttt{tree} does not have the attribute \texttt{offset}.  Instead the \texttt{mod} attribute should be added to the accumulated sum as follows:
\begin{verbatim}
modsum += tree.mod
\end{verbatim}

\onecolumngrid

\end{document}